\documentclass[11pt]{article}

\usepackage[preprint]{acl}
\usepackage{setspace} 
\usepackage{array}
\usepackage{longtable}
\usepackage{ltxtable} 
\usepackage{tabularx}   
\usepackage[table]{xcolor} 
\usepackage{array}          
\definecolor{highlightblue}{RGB}{219, 238, 246}

\definecolor{charfang}{RGB}{30, 144, 255} 
\definecolor{charchen}{RGB}{255, 105, 180} 
\definecolor{Plot}{RGB}{46, 139, 87} 
\usepackage{float}

\usepackage[T1]{fontenc}
\usepackage{lmodern}  

\usepackage{graphicx}

\usepackage{times}
\usepackage{latexsym}
\usepackage{cuted}
\usepackage[T1]{fontenc}
\usepackage{booktabs}   
\usepackage{tabularx}   
\usepackage{makecell}   
\usepackage{arydshln}   
\usepackage[utf8]{inputenc}
\usepackage{multirow}
\usepackage{adjustbox}

\usepackage{microtype}

\usepackage{inconsolata}

\usepackage{graphicx}

\usepackage{booktabs}

\usepackage{amsmath}
\usepackage{colortbl}
\usepackage{xcolor}

\usepackage{tcolorbox}

\usepackage{placeins}

\tcbuselibrary{breakable}
\usepackage{pifont}   
\usepackage{amssymb}  

\title{Beyond Direct Generation: A Decomposed Approach to Well-Crafted Screenwriting with LLMs}

\author{
    \textbf{Hang Lei\textsuperscript{1}}\thanks{Corresponding author.} \quad
    \textbf{Shengyi Zong\textsuperscript{1}}\quad
    \textbf{Zhaoyan Li\textsuperscript{1}} \quad
    \textbf{Ziren Zhou\textsuperscript{1,2}} \quad
    \textbf{Hao Liu\textsuperscript{1}}\quad
    \textbf{Liang Yu\textsuperscript{1}}
    \\[1ex] 
    \textsuperscript{1}Alibaba Group, \quad
    \textsuperscript{2}Peking University
    \\[1ex] 
    \texttt{\{leihang.lh,zongshengyi.zsy,lzy434483,zhouziren.zzr,lh414475,deyi.yl\}@alibaba-inc.com}
} 

\begin{document}
\maketitle
\renewcommand{\thefootnote}{\arabic{footnote}}
\setcounter{footnote}{-1}
\begin{abstract}
The screenplay serves as the foundation for television production, defining narrative structure, character development, and dialogue. While Large Language Models (LLMs) offer great potential in this creative process, direct end-to-end generation approaches often fail to produce well-crafted screenplays. We argue this failure stems from forcing a single model to simultaneously master two disparate capabilities: creative narrative construction and rigid format adherence.
To enable LLMs to generate high-quality screenplays, we introduce Dual-Stage Refinement (DSR), a decomposed framework that explicitly decouples creative narrative generation from format conversion. In the first stage, the framework transforms a brief outline into rich, novel-style prose. The second stage then refines this prose into a professionally formatted screenplay. This separation enables the model to specialize in one distinct capability at each stage.
A significant challenge in implementing DSR is the scarcity of paired outline-to-novel data. We address this through a hybrid data synthesis strategy that combines reverse synthesis (deconstructing existing screenplays into structured inputs) and forward synthesis (generating high-quality novel-style texts as training targets).
Extensive experiments show that in blind evaluations by professional screenwriters, screenplays generated by DSR achieve a $75\%$ win rate against strong baselines like Gemini-2.5-Pro and reach $82.7\%$ of human-level performance, demonstrating that decomposed generation architecture is highly effective for specializing LLMs in complex creative domains.
\end{abstract}

\section{Introduction}
The advent of LLMs brings new potential to screenplay generation\footnote{The terms `screenplay' and `script' in this work refer to scripts for episodic television, not feature films.}. LLMs can serve as creative assistants by handling time-consuming tasks such as generating plot variations and character biographies, allowing writers to focus on higher-level creative decisions. However, current general-purpose LLMs must be adapted to understand screenplay-specific narrative structures, maintain story coherence, and manage complex character development. The challenge is developing specialized AI tools that enable writers to produce higher-quality screenplays efficiently.

A screenplay differs fundamentally from a novel as a structured framework for visual storytelling. Rather than relying on prose descriptions or internal monologue, screenplays employ "showing, not telling" through precise audiovisual language—scene headings, concise action lines, and authentic dialogue. Consequently, an effective screenplay generation model must maintain narrative coherence, ensure character consistency, and adhere to strict formatting conventions.

Current research has made notable progress in controllable story generation, developing models that follow plot outlines~\citep{rashkin2020plotmachinesoutlineconditionedgenerationdynamic} or incorporate commonsense knowledge~\citep{wang2022incorporatingcommonsenseknowledgestory}. In screenwriting, systems like Dramatron~\citep{mirowski2022cowritingscreenplaystheatrescripts} employ hierarchical prompt chaining.

However, these approaches focus primarily on narrative content and logical coherence. We address a more complex challenge: generating production-ready screenplays that require both creative narrative generation ("what to write") and rigid format conversion ("how to write"). Handling both skills jointly increases training complexity and often produces outputs that superficially mimic screenplay style but lack professional quality. This suggests decoupling these skills, allowing the model to focus on one task at a time.

We introduce Dual Stage Refinement (DSR), a decomposed framework that separates screenplay generation into two stages. The first stage translates high-level outlines into rich, novel-style prose, focusing purely on storytelling. The second stage refines this prose into professionally formatted screenplays. This separation enables the model to master one distinct skill at each step.

Implementing DSR presents a data challenge: the lack of paired outline-to-novel data for training the first stage. We address this through a hybrid data synthesis strategy combining reverse synthesis (deconstructing screenplays into structured inputs) and forward synthesis (generating high-quality novel-style targets).

Blind evaluations by professional screenwriters demonstrate DSR's effectiveness: a $75\%$ win rate against SOTA models like Gemini-2.5-Pro and Claude-Sonnet-4, reaching $82.7\%$ of human-level performance.

Our main contributions are: \textbf{(1)} The DSR framework that decouples screenplay generation into creative narrative development and format conversion, addressing limitations of end-to-end approaches; \textbf{(2)} A hybrid data synthesis strategy combining reverse and forward synthesis to resolve data scarcity; \textbf{(3)} Comprehensive validation showing DSR significantly outperforms strong baselines in blind evaluations by professional screenwriters.

\section{Related Works}
\textbf{Screenplay Generation.}
Early screenplay generation relied on retrieval-based methods \citep{10.1145/3507356}. Modern LLM-based frameworks like Dramatron \citep{mirowski2022cowritingscreenplaystheatrescripts} and IBSEN \citep{han2024ibsendirectoractoragentcollaboration} enable interactive co-writing but require substantial human supervision. \citet{tian2024largelanguagemodelscapable} found that while LLMs excel at surface-level qualities, they fail at "narrative intelligence" in end-to-end generation due to the difficulty of simultaneously managing creative construction and structural constraints. Unlike collaborative frameworks, our work tackles autonomous, high-quality generation.

\textbf{Planning-based Narrative Generation.}
Plan-and-write approaches use sparse plans \citep{brei2023returning,wang2024guiding,wang2023improving} or detailed outlines \citep{yang2023docimprovinglongstory}, but these create semantic gaps between planning and execution. Sophisticated pipelines like CML-BENCH \citep{zheng2025cmlbenchframeworkevaluatingenhancing}, HoLLMwood \citep{kor2023hollmwood}, and R$^2$ \citep{lin2025r} still rely on lossy, non-narrative representations. Iterative refinement methods like Re3 \citep{yang2022re3generatinglongerstories} attempt costly post-generation fixes.

We address these limitations by using novels as dense, narratively-complete intermediate states, decoupling creative generation (outline-to-novel) from format conversion (novel-to-screenplay).

\textbf{Data Synthesis for Narrative Tasks.}
High-quality training data is scarce for screenplay generation. Prior work employs reverse synthesis \citep{you2023eipe,huang2023ex3,ahuja2024finding} to extract plans from texts, or forward synthesis \citep{yang2024rlcd,zhu2024end} using LLMs to generate training data. However, these focus on single-stage generation.

Our two-stage approach requires outline-to-novel pairs unavailable in existing datasets. We employ hybrid synthesis: reverse synthesis extracts structured inputs from screenplays, then forward synthesis generates novel-style narratives, producing the training pairs needed for our framework.

\section{Methodology}
\label{sec:methodology}
This section presents DSR, our proposed framework for generating well-crafted screenplays. We begin by formally defining the screenwriting task and examining the limitations of direct end-to-end generation (Section~\ref{sec:task_formulation}). We then introduce the DSR framework's two-stage architecture that decouples creative narrative generation from format conversion (Section~\ref{sec:DSR Framework}). 

\begin{figure*}[t]
    \centering
    \includegraphics[width=\textwidth]{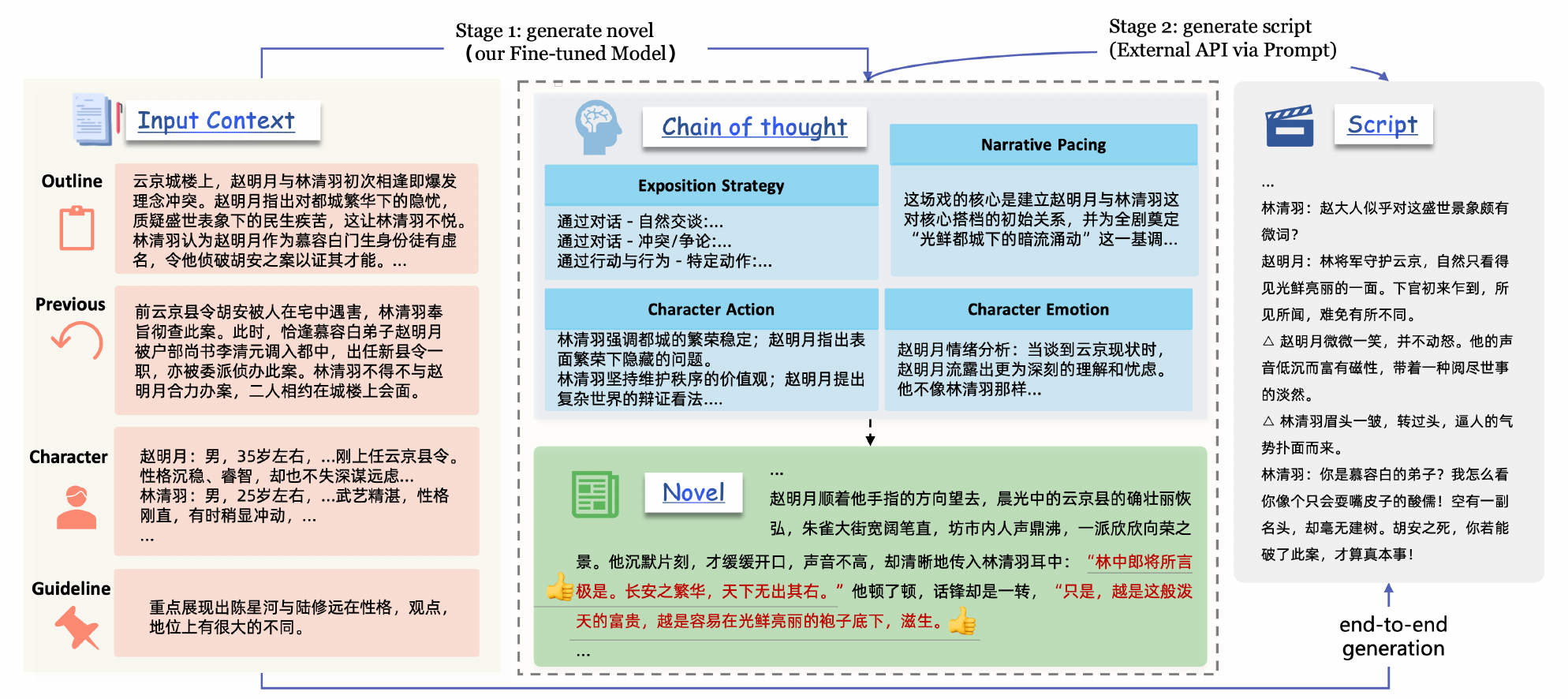} 
    \caption{The proposed two-stage generation pipeline. Stage 1 focuses on creative narrative generation (Outline-to-Novel), while Stage 2 handles structural formatting (Novel-to-Screenplay).}
    \label{fig:task_definition}
\end{figure*}

\subsection{Task Formulation}
\label{sec:task_formulation}
The primary objective is to generate a screenplay scene $S$, based on a set of structured inputs $X$. The input $X$ is a tuple $\{O, P, C, M\}$, where:
\begin{itemize}
    \itemsep0em
    \item Scene Outline $(O)$: The scene-by-scene outline, guiding the plot's progression.
    \item Previous Context $(P)$: The premise or prior context, ensuring narrative continuity.
    \item Character Profiles $(C)$: Character profiles, defining their backgrounds, personalities, and relationships.
    \item Metadata $(M)$: Specific instructions that guide the creative tone or content, such as "focus on the escalating conflict between Character A and B" or "this scene must end on a cliffhanger".
\end{itemize}

A fundamental challenge in this domain is the scarcity of paired training data. The available resources for our task consist solely of finalized screenplays from aired television series. We lack access to the corresponding outlines or authorial notes that writers originally used to create these screenplays. This data-availability constraint necessitates the construction of our own training pairs. Our primary task, therefore, is to create a high-quality dataset $\mathcal{D} = {\{(X_i, S_i)\}}_{i=1}^{|\mathcal{D}|}$, where the inputs $X_i$ are reverse-engineered to serve as plausible creative briefs for the existing screenplays $S_i$.

It is crucial to acknowledge that screenwriting is an open-ended creative task. For any given input $X_i$, there is no single "ground-truth" screenplay. Therefore, in our constructed dataset, the target screenplay $S_i$ should be viewed not as a definitive answer, but as a high-quality reference representing one point sampled from the vast space of possible valid solutions. This reference demonstrates the desired narrative structure, character voice, and screenplay format. The model's objective is not to replicate this specific sample, but to learn the general mapping from a creative brief to a professionally valid screenplay instance.

With this formulation and the constructed dataset, a standard Supervised Fine-Tuning (SFT) approach aims to train a model $M_\theta$ with parameters $\theta$ to directly learn the conditional probability distribution $P(S|X)$. The objective is to find the parameters $\theta^*$ that maximize the total log-likelihood of the target screenplays across the dataset $\mathcal{D}$. This is formally expressed as:

\begin{equation} \label{eq:sft_objective}
\theta^* = \operatornamewithlimits{arg\,max}_{\theta} \sum_{(X,S) \in \mathcal{D}} \log P(S | X; \theta)
\end{equation}

A straightforward approach to this task would be to train a model to directly generate the final screenplay from the input context in a single step. This naive baseline is visually represented by the blue arrow labeled "end-to-end generation" in Figure~\ref{fig:task_definition}, which depicts a direct mapping from the Input Context (comprising Outline, Previous scenes, Character profiles, and Guidelines) to the final Script, bypassing the intermediate Novel representation and the explicit Chain of Thought (CoT) reasoning process.
Our preliminary experiments with this end-to-end approach revealed suboptimal results. The generated screenplays often suffer from a lack of thematic focus, out-of-character dialogue, and insufficient character development. We attribute these shortcomings to the \textbf{Task Coupling Dilemma} faced by single-stage models. Specifically, the model is forced to simultaneously master two disparate skills: 
\noindent\textbf{(1) Narrative Generation:} The creative task of elaborating a story from a high-level outline into rich, detailed prose (as shown in the Novel box in Stage 1). This requires imaginative expansion to design the scene's pacing, its sequential development, and the intricate chain of cause and effect that drives the story forward. The model must reason through multiple dimensions simultaneously, including Exposition Strategy, Narrative Pacing, Character Action, and Character Emotion (illustrated in the Chain of Thought boxes), to construct a coherent dramatic skeleton.
\noindent\textbf{(2) Format Conversion:} The task of transforming descriptive narratives into the visual and auditory language of screenwriting (Stage 2 in Figure~\ref{fig:task_definition}). This means replacing narrative exposition with concrete visual actions and performable dialogue that convey the same story information through what can be seen and heard on screen, while following screenplay formatting conventions.

When training on dataset $\mathcal{D}$ to learn $P(S|X)$ in a single stage, gradients must simultaneously improve both narrative quality and format adherence, which can lead to conflicting optimization directions. Our proposed two-stage pipeline resolves this by introducing the Novel as an intermediate representation that decouples the learning objectives. Stage 1 optimizes for narrative generation independently, while Stage 2 optimizes for format conversion, eliminating gradient conflicts and simplifying each optimization problem.

\begin{figure*}[t]
    \centering
    \includegraphics[width=\textwidth]{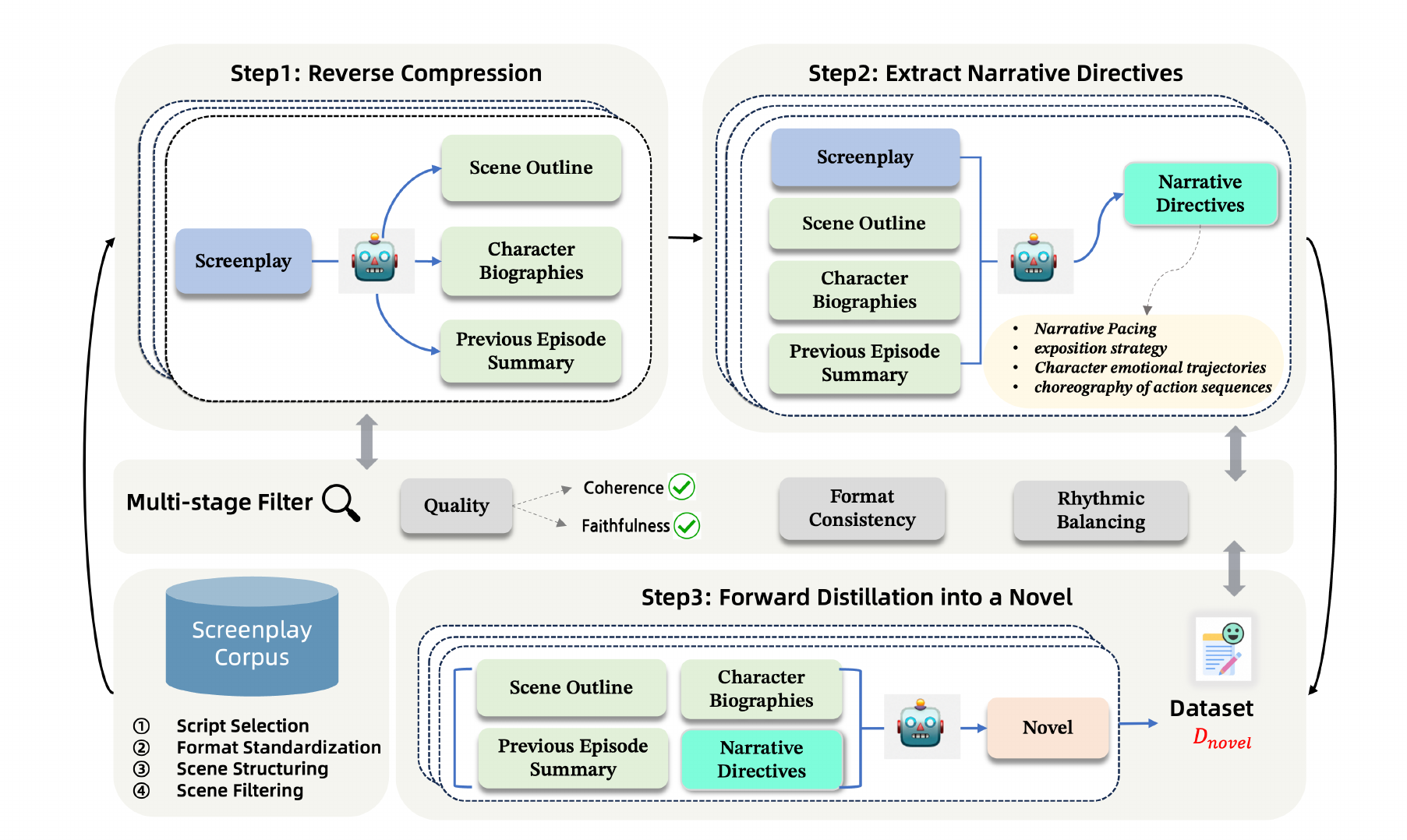} 
    \caption{Workflow of the data synthesizing process. The goal of Step1 is to extract the core structural and narrative elements from a complete screenplay. Step2 aims to understand the "why" behind the screenplay. The objective of Step3 is to generate a new, long-form narrative (a novel) based on the structured information and narrative directives obtained in the previous steps.}
    \label{fig:pipeline}
\end{figure*}

\subsection{The DSR Framework}
\label{sec:DSR Framework}
Following the rationale for a decomposed strategy, we now detail the two-stage design of our DSR framework. As illustrated in Figure~\ref{fig:task_definition}, this framework decouples creative narrative generation from stylistic format conversion by introducing novel-style prose as an intermediate representation, denoted as $N$. The complete generation process is formulated as a sequential sampling procedure:

\begin{description}
    \item[Stage 1: Outline-to-Novel Expansion with CoT.] The model learns to generate a Chain-of-Thought (CoT) analysis $I_{c}$, followed by the intermediate novel $N$, conditioned solely on the input $X$:
      \[
    (\hat{I}_{c}, \hat{N}) \sim P(I_{c}, N \mid X; \theta_1)
    \]
    
    \item[Stage 2: Novel-to-Screenplay Conversion.] The generated narrative prose $\hat{N}$ is then transformed into a structurally-correct screenplay $S$, guided by the original input $X$:
    \[
    \hat{S} \sim P(S \mid \hat{N}, X)
    \]
\end{description}

This decomposition corresponds to a probabilistic model where the target distribution $P(S|X)$ is implicitly modeled through marginalizing over the latent narrative representation $N$. 

The choice of novelistic prose as the intermediate representation is a central component of our framework's design. This choice is motivated by two observations about LLMs~\citep{openai2024gpt4technicalreport,touvron2023llamaopenefficientfoundation}. First, while LLMs are pre-trained on vast corpora of narrative texts such as novels and stories, they have seen comparatively limited well-formatted screenplay data during training. Second, there exists a significant difference in information density between the two formats: novels can unfold plots with extensive descriptions and details, whereas screenplays must compress the same narrative into a much more concise format with strict structural constraints.  Therefore, we introduce the novel as an intermediate representation, decomposing the generation task into two subtasks: first generating a novel from the outline, then converting the novel into a screenplay.

Notably, the intermediate novel is not a traditional literary novel. While a literary novel can describe characters' inner thoughts and use expressions like metaphors, the intermediate novel is specifically designed as a screenplay-oriented descriptive text. It serves as a practical guide for visual storytelling, with content constrained to observable actions and audible dialogue rather than abstract thoughts and emotions. This design bridges the gap between narrative generation and screenplay formatting. For instance, rather than stating a character's inner state as \textit{"He was consumed by regret"}, it describes a performable moment: \textit{"He stared at the cracked photograph, his jaw tight, before slowly closing his eyes"}.

The first stage \textbf{Outline-to-Novel Expansion} trains a model $M_{\theta_1}$ to perform both reasoning and creative generation. Its objective is to learn a mapping from the input brief $X$ to a structured output containing both a CoT analysis $I_c$ and the full novel prose $N$. By training the model to first generate a CoT, we encourage it to develop an internal reasoning process that enhances the quality and coherence of the subsequent novel generation. To achieve this, we train the model on a high-quality dataset $\mathcal{D}_{novel} = \{(X_i, (I_{c,i}, N_i))\}_{i=1}^{|\mathcal{D}_{novel}|}$.

Creating such a high-quality dataset is challenging. We have access to professionally written screenplays $S$, but not the original high-level briefs $X$ used to create them. We therefore develop a hybrid data synthesis strategy illustrated in Figure~\ref{fig:pipeline}, combining the strengths of reverse and forward synthesis to construct our training pairs. The input $X$ is created via reverse synthesis to resemble a realistic writer's brief, while the target $N$ is created via forward synthesis to ensure high fidelity and narrative richness.

The foundation of our synthesis strategy is a high-quality, pre-processed screenplay corpus.\footnote{See Appendix~\ref{app:data_pipeline} for detailed preprocessing steps.} As shown in the lower-left corner of Figure~\ref{fig:pipeline}, all raw scripts first pass through a standardized cleaning pipeline to ensure data quality and consistency. This resulting curated corpus contains over $200$ series with more than $50000$ scenes, serving as the source material for all subsequent synthesis tasks.

Based on this clean corpus, we perform the synthesis process in two main steps.

\textbf{Part A: Reverse Synthesis of Inputs ($X$) and Narrative Directives ($I_c$)} 
As depicted in Step 1: Reverse Compression of Figure~\ref{fig:pipeline}, our process begins by taking a professional screenplay $S$ from the pre-processed corpus and reverse-engineering the corresponding input $X$ to simulate a writer's outline. However, reverse synthesis presents several key challenges. A primary difficulty lies in achieving the correct level of detail: the input must provide enough critical information (\textit{e.g.}, character conflicts and plot turning points) without including so much that the task becomes unrealistic. Moreover, reliance on LLMs (\textit{e.g.}, GPT-4) introduces the risk of hallucinations that create details inconsistent with the original screenplay, thereby reducing the quality of training data.

To overcome these difficulties, we reframe the task from simple "summarization" to "creative intent reconstruction" through iterative prompt engineering, focusing on extracting what happens in the story. Furthermore, all outputs undergo a rigorous human-in-the-loop review pipeline involving manual editing and quality filtering (Multi-stage Filter in Figure~\ref{fig:pipeline}). The review process guarantees that the final input $X$ is accurate, representative of a realistic writer's outline, and maintains both coherence and faithfulness to the source material.

Concurrently, we move beyond the plot-level details of $X$ to capture the latent authorial strategy: the choices that dictate how the story is told. As shown in Step 2: Extract Narrative Directives of Figure~\ref{fig:pipeline}, we analyze the source screenplay $S$ again to extract a set of parameters termed \textbf{Narrative Directives ($I_c$)}. These directives capture the underlying mechanics of storytelling, such as narrative pacing, the trajectory of character emotions, the choreography of action sequences, and the information disclosure strategy. This step provides a deeper layer of guidance that complements the structural information in $X$.

\textbf{Part B: Enhanced Forward Synthesis of Targets ($N$)} 
The goal of the forward synthesis phase is to generate our final training target: the novel $N$. While a forward approach guarantees consistency with its inputs, naive forward synthesis that conditions only on the structured input $X$ ($X \rightarrow N$) often produces narratives that are logical but lack creative depth and professional quality. We therefore develop an enhanced forward synthesis process. As shown in Step 3: Forward Distillation into a Novel of Figure~\ref{fig:pipeline}, we use a powerful teacher model (\textit{e.g.}, Gemini, GPT-4) to generate the target novel $N$. Crucially, the generation is conditioned on both the structural input $X$ from Step 1 and the narrative directives $I_c$ from Step 2:
\begin{equation}
\hat{N} \sim M_{teacher}(X, I_c)
\end{equation}
Here, $M_{teacher}$ denotes the "teacher model", a term from the teacher-student paradigm in machine learning, where a larger, highly capable model generates high-quality training data for a smaller, more specialized "student model". This enhanced process produces a target novel $N$ that is not only consistent with its input $X$ but also demonstrates narrative sophistication and design quality inspired by professional writing.

By combining reverse and forward synthesis, our final training pairs are structured as $(X_i, (I_{c,i}, N_i))$. The model is trained to sequentially generate the narrative directives $I_{c,i}$ followed by the novel $N_i$, conditioned on the input $X_i$. This design allows the model to learn both the reasoning process and the creative writing process.

With the high-quality dataset $\mathcal{D}_{novel}$, we apply supervised fine-tuning (SFT) to obtain $M_{\theta_1}$ from a base model. The training objective minimizes the standard negative log-likelihood loss over the entire target sequence, which is the concatenation of the narrative directives and the novel. Let $Y_i = (I_{c,i}, N_i)$ represent this full target sequence, where $Y_i = (y_{i,1}, \dots, y_{i,T_i})$. The loss function is:

\begin{equation}
\label{eq:sft_loss_final}
\begin{split}
\mathcal{L}_{SFT}(\theta_1) &= - \frac{1}{|\mathcal{D}_{novel}|} \sum_{(X_i,Y_i)\in\mathcal{D}_{novel}} \\
& \qquad \log P(Y_i | X_i; \theta_1) \\
&= - \frac{1}{|\mathcal{D}_{novel}|} \sum_{(X_i,Y_i)\in\mathcal{D}_{novel}} \\
& \qquad \sum_{t=1}^{T_i} \log P(y_{i,t} | y_{i,<t}, X_i; \theta_1)
\end{split}
\end{equation}

Upon generating the rich narrative prose $N$ from Stage 1, the second stage, \textbf{Novel-to-Screenplay Conversion}, performs the stylistic transformation. This is an inference-only stage that requires no additional fine-tuning. We leverage the powerful in-context learning capabilities of a separate large-scale model, denoted as $M_{api}$ (\textit{e.g.}, GPT-4), to approximate the distribution $P(S|N,X)$. The conversion is guided by a carefully engineered prompt $\pi$, which instructs the model to act as a professional screenwriter and provides clear formatting rules. The final screenplay $S$ is generated by sampling from this model:
\begin{equation}
S \sim M_{api}(\pi(N, X))
\end{equation}

The prompt $\pi(N, X)$ is carefully designed to provide clear context and instructions. It begins by establishing a specific persona through a role-playing directive, such as "You are a professional screenwriter," followed by a clear task definition. To ensure structural correctness, the prompt includes explicit formatting rules with examples of proper scene headings, action lines, and dialogue. Finally, the novelistic prose $N$ generated in Stage 1 is appended to these instructions as the input text for conversion.

In summary, our two-stage approach systematically develops both creative aspects of screenwriting. Stage 1 focuses on imaginative plot design and narrative development, while Stage 2 refines the output into proper screenplay format with appropriate character expression. By separating these tasks, we enable the model to produce scripts with both structural integrity and narrative quality.

\begin{figure}[t]
    \centering
    \includegraphics[width=\columnwidth]{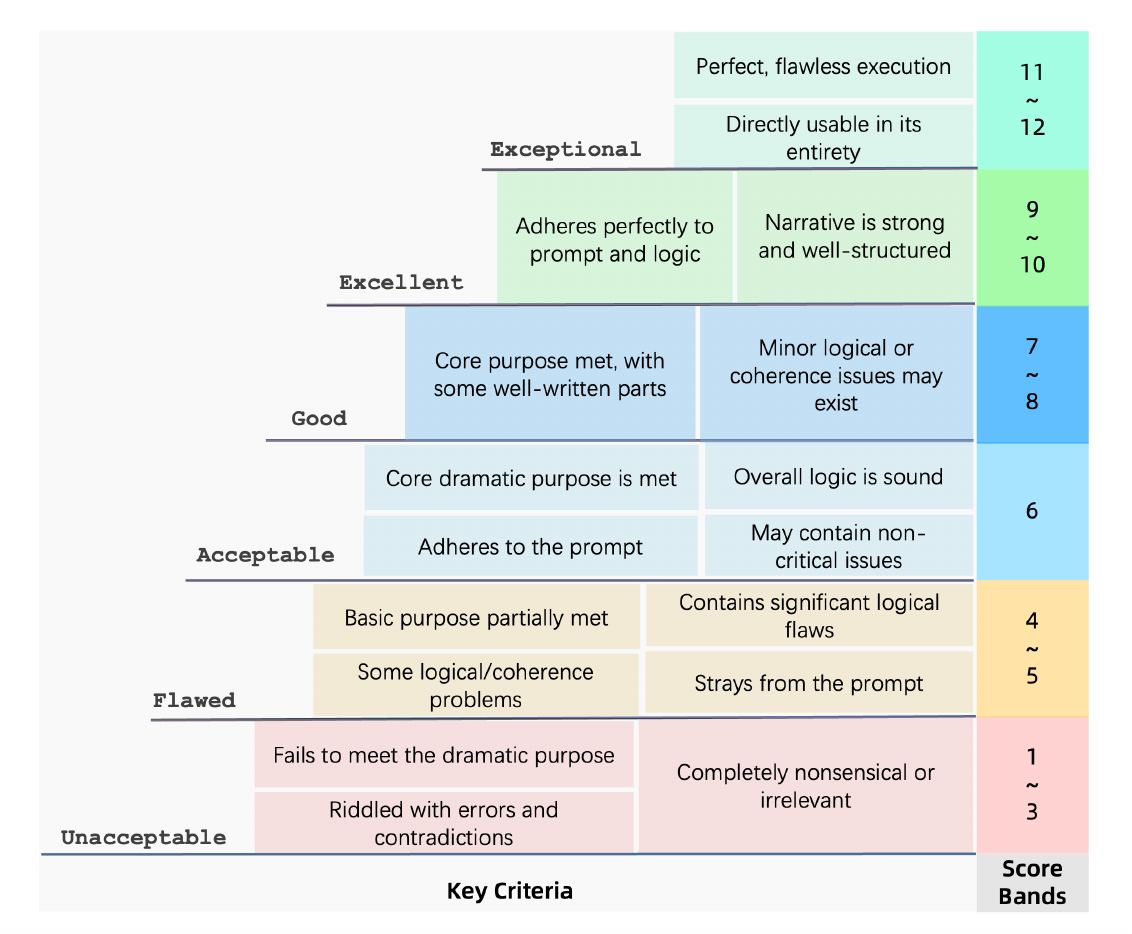}
    \caption{Holistic quality scoring rubric for screenplay evaluation. Each tier represents a quality level with corresponding score ranges and performance criteria.}
    \label{fig:Evaluation}
\end{figure}

\section{Experiments}
This section details the datasets, evaluation metrics, models, and implementation specifics used to validate our proposed two-stage methodology for screenplay generation.

\subsection{Experimental Setting}
\textbf{Datasets.} Our experiments utilize two primary datasets: a large-scale dataset for fine-tuning and a custom high-quality dataset for evaluation.\footnote{See Appendix~\ref{app:dataset_details} for detailed dataset statistics and composition.}
\begin{itemize}
    \item \textit{\textbf{Fine-Tuning Dataset}}: We constructed a comprehensive fine-tuning dataset comprising $50,000$ samples covering diverse genres.
    \item \textit{\textbf{Evaluation Dataset}}: We built a high-quality test set tailored for screenplay evaluation, comprising $32$ distinct scenes sourced from four different television series, carefully curated by human experts.
\end{itemize}

\textbf{Evaluation Metrics.} Given the creative and subjective nature of screenplay writing, our evaluation relies on expert human judgment, supplemented by comparative and diagnostic metrics.\footnote{See Appendix~\ref{app:evaluation_metrics} for complete metric definitions and scoring rubric details.}

The primary evaluation metric is a holistic quality score assigned by professional Chinese television drama screenwriters. We involved over $20$ experienced screenwriters in the evaluation, with each screenplay independently scored by all of them. The final score for each screenplay is the average of all screenwriters' ratings. The scoring rubric, illustrated in Figure~\ref{fig:Evaluation}, employs a 12-point scale organized into six quality tiers.

In addition, we designed several auxiliary metrics including Variance, Error Counts, Ratio to Human, and Win Rate to enable a more comprehensive analysis of model performance.

\textbf{Models for Comparison.} The experimental design encompasses a diverse set of models for comprehensive evaluation, including both fine-tuned models and proprietary state-of-the-art APIs.\footnote{See Appendix~\ref{app:model_details} for complete model specifications.}

\begin{itemize}
    \item \textit{\textbf{Fine-Tuned Models}}: Several models of varying scales and specializations are compared, including \texttt{Generic-LLM} (\texttt{Qwen-14B-Chat} and \texttt{QwQ-32B}), \texttt{Qwen-72B-Chat}, and \texttt{Qwen-72B-CPT}.
    \item \textit{\textbf{API Models}}: The best-performing fine-tuned model is evaluated against \texttt{Claude-Sonnet-4} and \texttt{Gemini-2.5-Pro}.
\end{itemize}

\textbf{Implementation Details.} To ensure a fair and rigorous comparison, all fine-tuned models were trained on the same dataset using identical hyperparameters.\footnote{See Appendix~\ref{app:implementation} for complete hyperparameters and hardware specifications.}

\subsection{Main Results}
Table~\ref{tab:Evaluation} presents the evaluation results. Compared to the base model \texttt{Qwen-72B-Chat} and state-of-the-art LLMs \texttt{Gemini-2.5-Pro} and \texttt{Claude-Sonnet-4}, screenplays generated by our DSR framework achieve a higher win rate, lower variance, and attain the highest average score of $8.06$. Although this remains below the human-written reference score of $9.75$, reaching approximately 83\% of professional quality, the generated scripts are sufficiently high-quality to serve as viable first drafts for further refinement by professional writers.

\begin{figure}[t]
    \centering
    \includegraphics[width=\columnwidth]{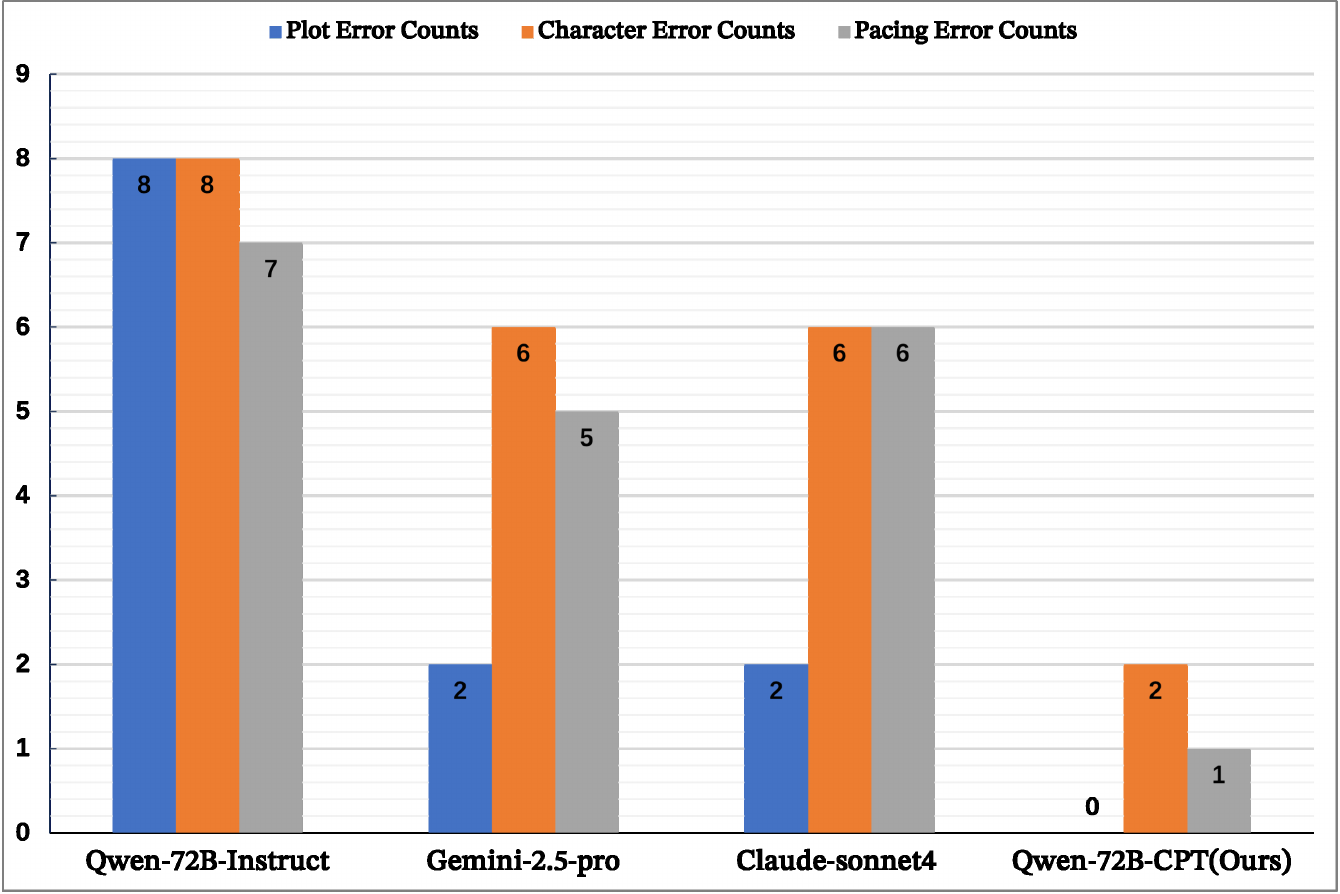}
    \caption{Error frequency comparison across models.}
    \label{fig:Error Analysis Across Models}
\end{figure}

Furthermore, Figure~\ref{fig:Error Analysis Across Models} illustrates the frequency of different error types across various models. Our DSR framework substantially reduces error rates compared to \texttt{Gemini-2.5-Pro}, particularly in character development and narrative pacing. This indicates that DSR strengthens the model's capacity to construct consistent and psychologically coherent characters, while also enabling the generation of scripts with better narrative pacing. Additionally, the DSR approach demonstrates superior controllability in text generation, exhibiting minimal deviation from the input outline, which suggests stronger alignment between the generated content and the intended narrative structure.

These results show that our method enhances LLMs' performance in screenplay generation, particularly in managing narrative structure, character development, and dramatic expression. This advancement represents a meaningful step toward practical AI-assisted screenwriting, bringing automated text generation closer to professional storytelling standards.

\begin{table*}[t]
\centering
\caption{Evaluation results of different models on screenplay generation. The \textit{Expert Score} is averaged across ratings from over $20$ professional screenwriters. Red superscripts with upward arrows indicate improvement over the baseline (\texttt{Qwen-72B-Chat}). Bold numbers indicate the best performance among models in each metric.}
\label{tab:Evaluation}
\begin{tabular}{lccccc}
  \toprule
  \textbf{Model} & \textbf{Method} & \textbf{Expert Score} & \textbf{Variance} & \textbf{Ratio to Human} (\%) & \textbf{Win Rate} (\%) \\ 
  \midrule
  \texttt{Qwen-72B-Chat} & Prompt & 3.43 & 0.19 & 35.18 & 0.0 \\
  \texttt{Claude-Sonnet-4} & Prompt & 6.69\textsuperscript{\textcolor{red}{$\uparrow$3.26}} & 0.41 & 68.61 & 12.5 \\ 
  \texttt{Gemini-2.5-Pro} & Prompt & 6.95\textsuperscript{\textcolor{red}{$\uparrow$3.52}} & 0.34 & 71.28 & 12.5 \\
  \midrule
  \rowcolor{highlightblue}
  \texttt{Qwen-72B-CPT} & DSR (Ours) & \textbf{8.06}\textsuperscript{\textcolor{red}{$\uparrow$4.63}} & \textbf{0.14} & \textbf{82.67} & \textbf{75.0} \\
  \midrule
  Human & - & 9.75 & 0.08 & - & -\\ 
  \bottomrule
\end{tabular}
\end{table*}

\subsection{Ablation Study}
To validate the design choices of our DSR framework, we conduct comprehensive ablation studies.\footnote{See Appendix~\ref{app:ablation_study} for complete ablation results and detailed analysis.} The key findings are: (1) Model scale is crucial—larger models significantly outperform smaller variants; (2) The two-stage DSR pipeline yields approximately $2.4$ points improvement over end-to-end generation; (3) Both continual pre-training (CPT) and Chain-of-Thought (CoT) reasoning contribute to performance gains; (4) Our Hybrid data synthesis strategy outperforms the Reverse-only approach, achieving the lowest variance of $0.14$. These findings collectively highlight the importance of appropriate model scale, domain-adaptive pre-training, and structured data synthesis in advancing LLMs for sophisticated creative applications.

\section{Conclusions}
In this paper, we focus on enabling large language models to generate high-quality screenplays from basic settings such as outlines and character profiles. Direct end-to-end generation is inadequate for this complex task: the generated screenplays may follow stylistic conventions, but often lack the deep structural integrity and storytelling substance required for professional use. Therefore, we propose the Dual-Stage Refinement (DSR) framework, which decomposes the single complex task into two distinct stages: creative narrative generation and strict format conversion. This decomposition allows models to focus on one specific skill at each stage, avoiding the poor performance caused by task entanglement. Implementing this framework requires paired training data for the narrative generation stage, which we obtain through an innovative hybrid data synthesis strategy.
The efficacy of the combined DSR framework and data synthesis strategy was validated through extensive experiments. In blind evaluations conducted by professional screenwriters, screenplays generated by the DSR framework achieved a $75\%$ win rate against strong baselines including Gemini-2.5-Pro, and reached $82.7\%$ of human-level performance. These results demonstrate that a decomposed generation framework enabled by tailored hybrid data synthesis significantly outperforms approaches that rely solely on prompt engineering to guide large, general-purpose language models. As screenplay generation is representative of complex creative writing tasks, this decomposition-based approach is likely applicable to broader creative content generation scenarios.

\section*{Limitations}
While our framework has proven effective, its current implementation has certain limitations. The data synthesis strategy relies on an initial corpus of high-quality human-written screenplays, which may limit its scalability to domains where such data is scarce. Future work will explore methods to reduce this dependency and enhance the scalability of the data synthesis process. Additionally, we plan to extend the DSR paradigm to other structured creative writing tasks, such as composing structured poetry or developing long-form fictional narratives, to further validate its generalizability across diverse creative domains.

\bibliography{custom}

\appendix

\section{DSR Framework Implementation Details}
\label{app:framework_details}

\subsection{Data Preprocessing Pipeline}
\label{app:data_pipeline}

The foundation of our synthesis strategy is a high-quality, pre-processed screenplay corpus. As shown in the lower-left corner of Figure~\ref{fig:pipeline}, all raw scripts first pass through a standardized cleaning pipeline to ensure data quality and consistency. The pipeline includes four main steps: \textbf{1) Script Selection} filters for relevant screenplays, \textbf{2) Format Standardization} unifies diverse writing styles into a single schema, \textbf{3) Scene Structuring} parses scenes into a consistent data format, and \textbf{4) Scene Filtering} removes noisy or irrelevant scenes.

\section{Experimental Details}
\label{app:experimental_details}

\subsection{Dataset Details}
\label{app:dataset_details}

\subsubsection{Fine-Tuning Dataset}
We constructed a comprehensive fine-tuning dataset comprising $50,000$ samples covering diverse genres, such as historical costume drama, fantasy/cultivation, espionage thrillers, and contemporary urban stories. Each sample consists of structured input (outline, previous events, character profiles, etc.) paired with the target output (narrative directives and novel for Stage 1 training, or screenplay for end-to-end training).

\subsubsection{Evaluation Dataset}
Current LLM benchmarks primarily assess general reasoning and instruction-following capabilities, lacking focus on complex creative tasks such as screenplay generation. We then built a high-quality test set tailored for screenplay evaluation. It comprises $32$ distinct scenes sourced from four different television series, carefully curated by human experts to evaluate plot coherence, thematic relevance, and character development.

\subsection{Evaluation Metrics Details}
\label{app:evaluation_metrics}

\subsubsection{Primary Metric: Holistic Quality Score}
The primary evaluation metric is a holistic quality score assigned by professional Chinese television drama screenwriters. We involved over $20$ experienced screenwriters in the evaluation, with each screenplay independently scored by all of them. The final score for each screenplay is the average of all screenwriters' ratings. The scoring rubric, illustrated in Figure~\ref{fig:Evaluation}, employs a 12-point scale organized into six quality tiers: "Unacceptable" (1\textasciitilde3), "Flawed" (4\textasciitilde5), "Acceptable" (6\textasciitilde7), "Good" (8), "Excellent" (9\textasciitilde10), and "Exceptional" (11\textasciitilde12). Each tier is characterized by specific performance levels across key criteria including adherence to the prompt, narrative structure, logical coherence, and fulfillment of dramatic purpose. At the lower end, "Unacceptable" and "Flawed" scripts fail to meet dramatic requirements, contain significant logical errors, or lack coherence. The "Acceptable" tier (6\textasciitilde7) marks the quality threshold where scripts successfully meet core dramatic requirements despite potential minor issues. "Excellent" scripts (9\textasciitilde10) demonstrate perfect adherence to the prompt and logic with strong narrative structures, while "Exceptional" scripts (11\textasciitilde12) exhibit flawless execution and are directly usable.

\subsubsection{Auxiliary Metrics}
\begin{itemize}
    \item \textit{\textbf{Variance}}: The variance of expert scores for each model's outputs reflects the stability and consistency of screenplay generation. Lower variance indicates more consistent quality across generated scripts, demonstrating greater reliability in creative performance.
    
    \item \textit{\textbf{Error Counts}}: Expert annotators identified and categorized specific errors in each generated script for fine-grained evaluation of model behavior. Three error categories were defined: (1) \textit{plot coherence}, referring to departure from the input outline or narrative intent; (2) \textit{character development}, including inadequate or incorrect character portrayal with misaligned motivations or behaviors; and (3) \textit{narrative pacing}, where plot development is either overly brief or excessively drawn out. The frequency and distribution of these errors across models reveal their respective weaknesses and failure modes.
    
    \item \textit{\textbf{Ratio to Human}}: This metric computes the ratio between each model's average score and the average score of professionally written reference scripts, quantifying how close generated screenplays are to human-level quality. The normalized measure indicates the extent to which a model approaches human-level storytelling proficiency.
    
    \item \textit{\textbf{Win Rate}}: In pairwise comparative evaluations, expert evaluators directly compared model outputs under identical conditions. The win rate is defined as the percentage of times a model's output was selected as the best in head-to-head comparisons. This preference-based metric captures qualitative differences that may not be fully reflected in absolute scores.
\end{itemize}

\subsection{Model Specifications}
\label{app:model_details}

\subsubsection{Fine-Tuned Models}
The core of this investigation involves models fine-tuned on the custom dataset. Several models of varying scales and specializations are compared to analyze their impact on screenplay generation. As baselines, general-purpose chat models of different scales are included: the smaller-scale models \texttt{Generic-LLM} (\texttt{Qwen-14B-Chat} and \texttt{QwQ-32B}), and the larger \texttt{Qwen-72B-Chat}. The primary focus of this comparison is \texttt{Qwen-72B-CPT}, which was created by performing continual pre-training (CPT) on \texttt{Qwen-72B}. The CPT utilized a narrative corpus of approximately \textbf{30 billion tokens}, curated from high-quality novels and professional screenplays.

\subsubsection{API Models}
For comparison with state-of-the-art API models, the best-performing fine-tuned model is evaluated against \texttt{Claude-Sonnet-4} and \texttt{Gemini-2.5-Pro} using a carefully crafted prompt that follows the same structured input format.

\subsection{Implementation Details}
\label{app:implementation}

\subsubsection{Training Configuration}
To ensure a fair and rigorous comparison, all fine-tuned models (\texttt{Generic-LLM}, \texttt{Qwen-72B-Chat}, and \texttt{Qwen-72B-CPT}) were trained on the same dataset using identical hyperparameters. Specifically, a cosine learning rate scheduler was employed with a peak learning rate of $5 \times 10^{-6}$ and a $10\%$ warmup phase. All models were trained for one epoch with a global batch size of $32$. The fine-tuning process was conducted on a cluster of 8$\times$ NVIDIA A100 (80GB) GPUs.

\subsubsection{Inference Configuration}
For inference, consistency was maintained across all models to ensure comparable outputs. All models used the same decoding strategy with a temperature of $0.7$ and a \textit{top-p} value of $0.9$ to balance creative diversity and output coherence.

\subsection{Ablation Study: Complete Results and Analysis}
\label{app:ablation_study}

To investigate the impact of different base models and data synthesis strategies on final performance, we conduct a comprehensive ablation study, with results presented in Table~\ref{tab:ablation_full}.

\begin{table*}[h]
\centering
\caption{Complete ablation study of the DSR framework. We evaluate the impact of key components including model scale, continual pre-training (CPT), task-decoupled pipeline, and data synthesis strategies. The best score is in \textbf{bold}, and the second-best is \underline{underlined}.}
\label{tab:ablation_full}
\begin{tabular}{llcccc}
  \toprule
  \textbf{Pipeline} & \textbf{Model} & \textbf{Data Synthesis} & \textbf{CoT} & \textbf{Expert Score} & \textbf{Variance}\\ 
  \midrule
  \multirow{5}{*}{End-to-End}
  & \texttt{Qwen-14B-Chat} & - & w/o & 2.33 & 0.15 \\
  \cmidrule(lr){2-6}
  & \texttt{QwQ-32B} & - & w/o & 3.78 & 0.38 \\
  \cmidrule(lr){2-6}
  & \texttt{Qwen-72B-Chat} & - & w/o & 4.01 & 0.25 \\ 
  \cmidrule(lr){2-6}
  & \texttt{Qwen-72B-CPT} & - & w/o & 4.63 & 0.31 \\ 
  \cmidrule(lr){2-6}
  & \texttt{Qwen-72B-CPT} & - & w/ & 5.13 & 0.30 \\ 
  \midrule
  \multirow{4}{*}{DSR}
  & \texttt{Qwen-72B-Chat} & Reverse-only & w & 6.41 & 0.31 \\
  \cmidrule(lr){2-6}
  & \texttt{Qwen-72B-CPT} & Reverse-only & w & \underline{7.14} & 0.28 \\ 
  \cmidrule(lr){2-6}
  & \texttt{Qwen-72B-CPT} & Hybrid & w/o & 7.08 & 0.17 \\ 
  \cmidrule(lr){2-6}
  \rowcolor{highlightblue}
  & \texttt{Qwen-72B-CPT} & Hybrid & w/ & \textbf{8.06} & 0.14 \\ 
  \bottomrule
\end{tabular}
\end{table*}

\subsubsection{Impact of Model Scale}

First, unsurprisingly, model scale plays a crucial role in performance. The smaller variants \texttt{Qwen-14B-Chat} and \texttt{QwQ-32B} exhibit significantly weaker performance compared to \texttt{Qwen-72B-Chat}, confirming that larger model size is necessary for complex creative tasks like screenplay generation, which require deep narrative understanding and expressive language generation.

\subsubsection{DSR Pipeline vs. End-to-End Generation}

Second, the end-to-end generation approach performs worse than the two-stage DSR pipeline, with the latter yielding an improvement of approximately $2.4$ points. This performance gap validates the effectiveness of explicitly separating narrative generation and format conversion, allowing the model to focus on each task independently.

\subsubsection{Impact of Continual Pre-training and CoT}

Third, both continual pre-training and CoT reasoning contribute to performance improvements. CPT consistently boosts scores for both end-to-end and DSR pipelines, while incorporating CoT reasoning brings further gains. For instance, applying both CPT and CoT with the DSR pipeline achieves a score of $8.06$, compared to $4.01$ for the baseline model.

\subsubsection{Data Synthesis Strategy Comparison}

Finally, our Hybrid data synthesis strategy outperforms the Reverse-only strategy, achieving the lowest variance of $0.14$, which indicates more stable and consistent output quality. In the Reverse-only approach, both the training input and output are derived from reverse-engineering the screenplay. Specifically, the training input $X$ is constructed through reverse compression as described in Part A of Section~\ref{sec:DSR Framework}. The training output is also reverse-engineered: the novel is generated by converting the screenplay's dialogue and character actions into novelistic narrative prose. Our analysis reveals two key limitations of this approach. On the one hand, the converted novels retain the high information density of the original screenplays, resulting in comparable learning difficulty to end-to-end screenplay generation. On the other hand, the dual reverse-engineering process leads to input-output misalignment: key elements such as specific props or minor characters appearing in the converted novel may be absent from the independently reverse-compressed input outline. Models trained on such misaligned pairs learn to generate details ungrounded in their inputs, producing off-topic content during inference. In contrast, our Hybrid strategy employs forward synthesis for output generation, where a teacher model expands the reverse-compressed input into a novel. This ensures input-output consistency while demonstrating proper narrative expansion from concise specifications to detailed prose, enabling the model to learn more reliable and controllable generation patterns.

\section{Prompts for Data Synthesis}
\label{sec:appendix_prompts}
The prompts for our data synthesis strategy are organized according to its two main phases: Reverse Synthesis and Forward Synthesis. The initial prompts deconstruct a source screenplay ($S$) into a structured input ($X$) and a set of Narrative Directives ($I_c$). The final prompt then utilizes these components to guide the synthesis of the target novel ($N$). Note that for the prompts presented in the following figures, content enclosed in curly braces \{\} serves as a placeholder for dynamic inputs that vary with each sample. Furthermore, it should be clarified that while English translations are provided, all experiments were conducted exclusively with the original Chinese versions.

\subsection{Prompts for Structured Input (\textit{X})}
\begin{figure*}[t]
    \centering
    \includegraphics[width=\textwidth]{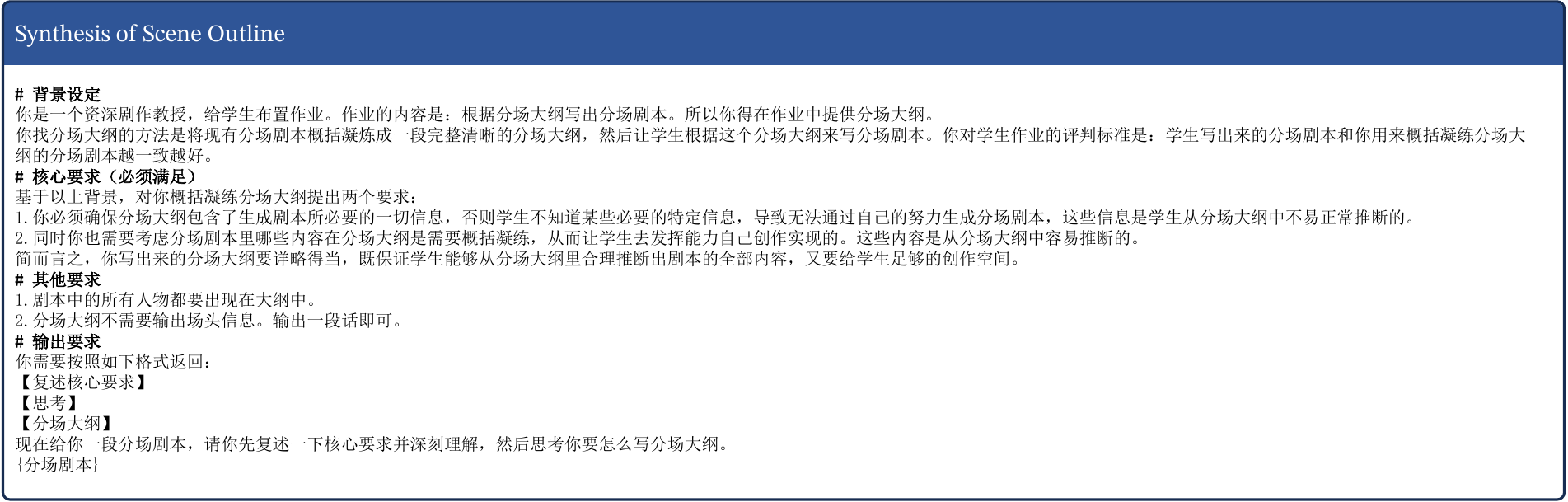} 
    \caption{The Prompt Template for Reverse Synthesis of Scene Outline.}
    \label{fig:Reverse Synthesis of Scene Outlines}
\end{figure*}

\begin{figure*}[t]
    \centering
    \includegraphics[width=\textwidth]{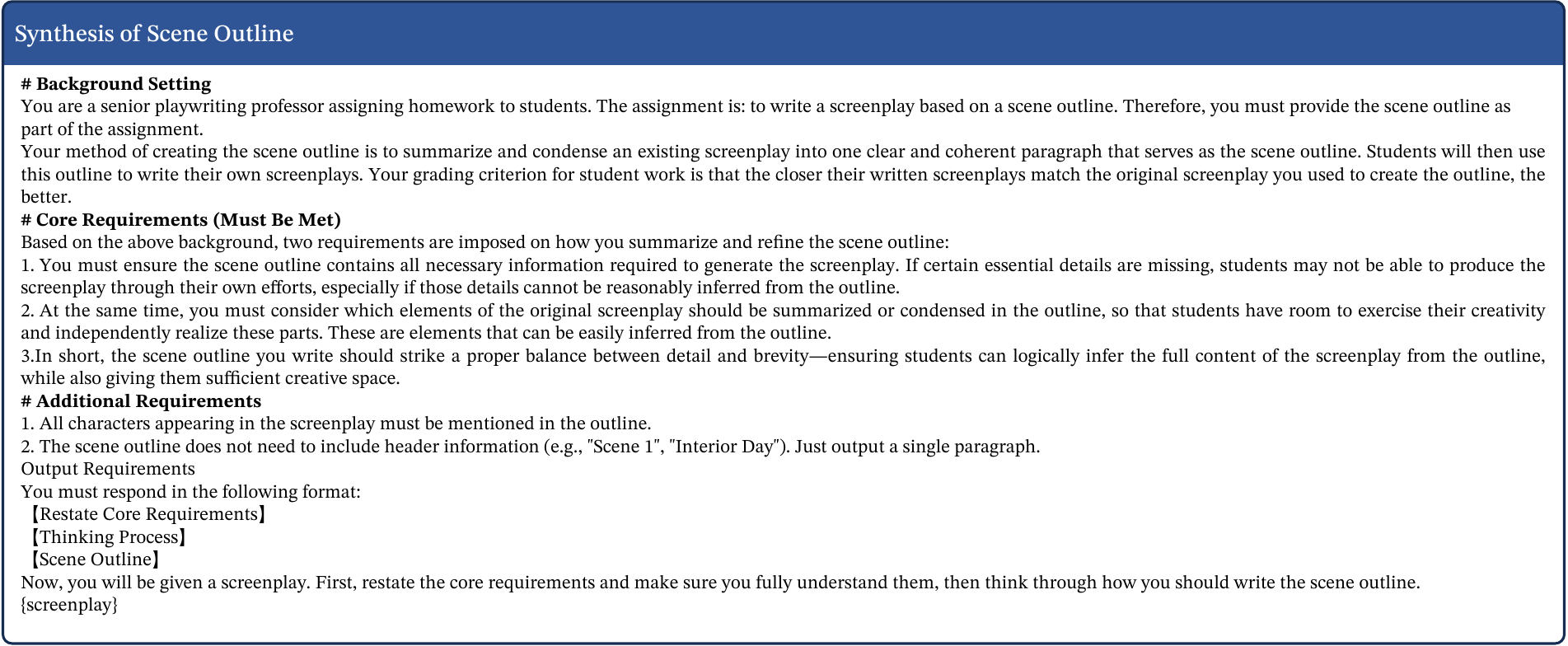} 
    \caption{The Prompt Template for Reverse Synthesis of Scene Outline translated into English.}
    \label{fig:Reverse Synthesis of Scene Outlines en}
\end{figure*}

\begin{figure*}[t]
    \centering
    \includegraphics[width=\textwidth]{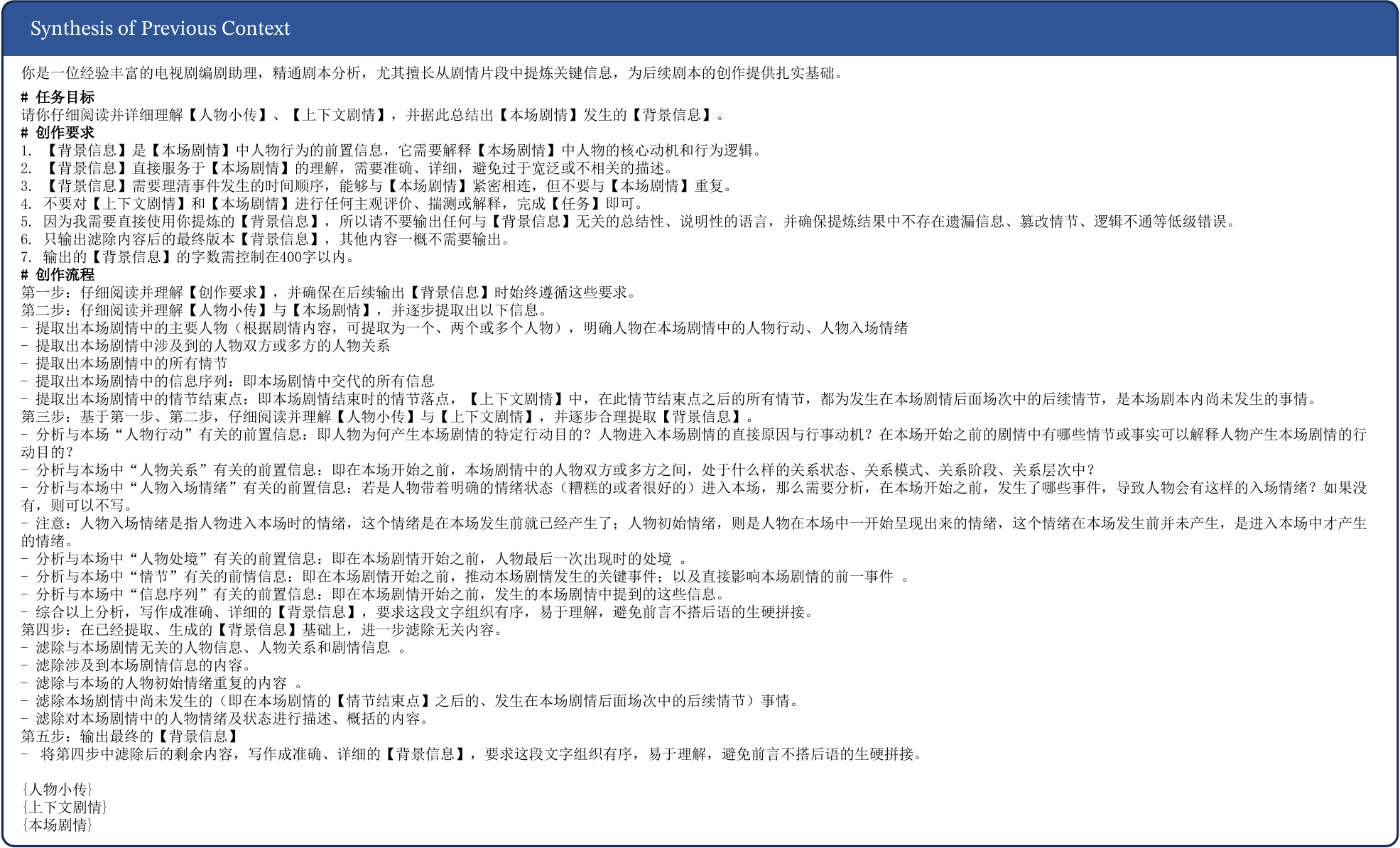} 
    \caption{The Prompt Template for Reverse Synthesis of Previous Context.}
    \label{fig:Reverse Synthesis of Previous Context}
\end{figure*}

\begin{figure*}[t]
    \centering
    \includegraphics[width=\textwidth]{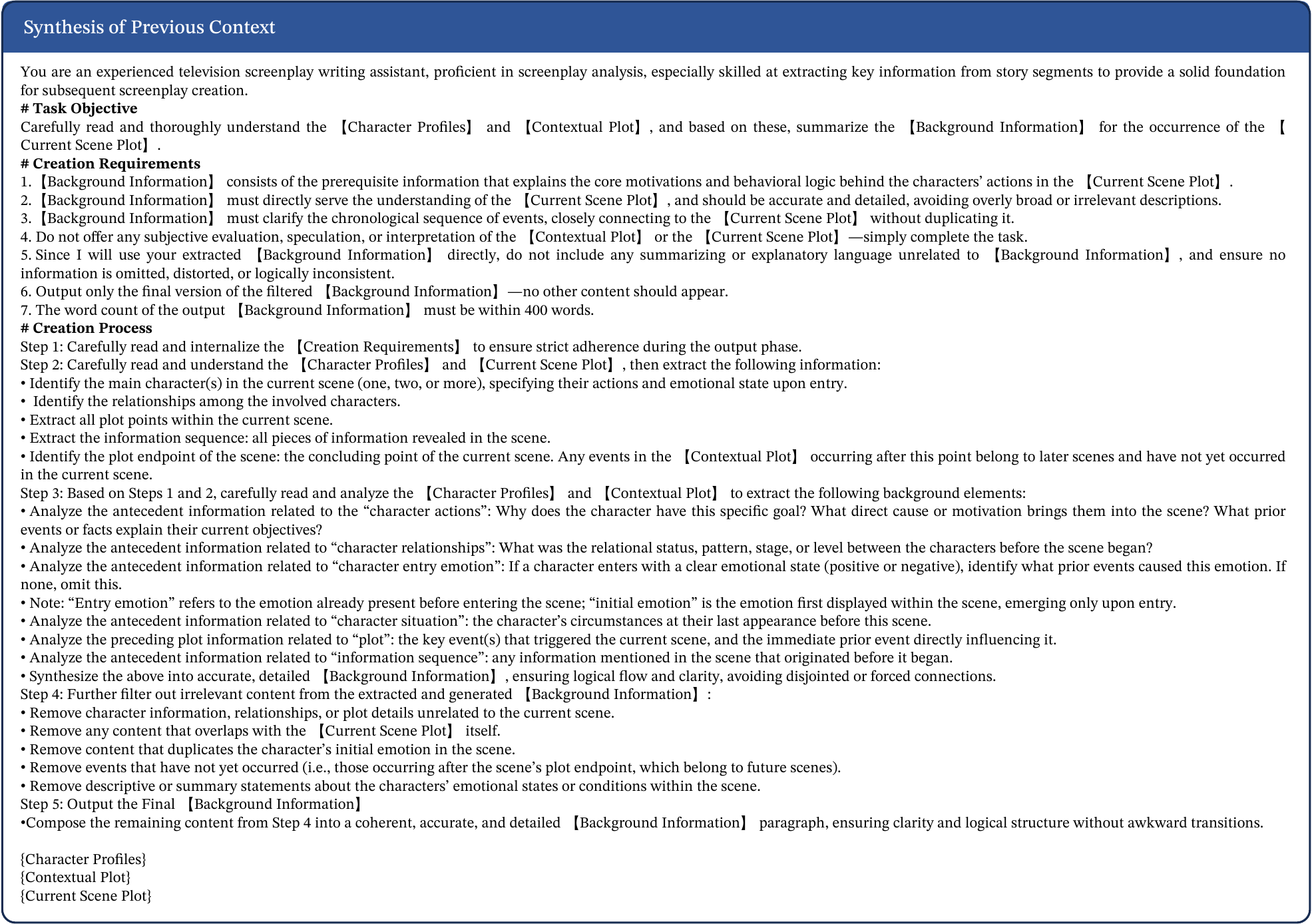} 
    \caption{The Prompt Template for Reverse Synthesis of Previous Context translated into English.}
    \label{fig:Reverse Synthesis of Previous Context en}
\end{figure*}

\begin{figure*}[t]
    \centering
    \includegraphics[width=\textwidth]{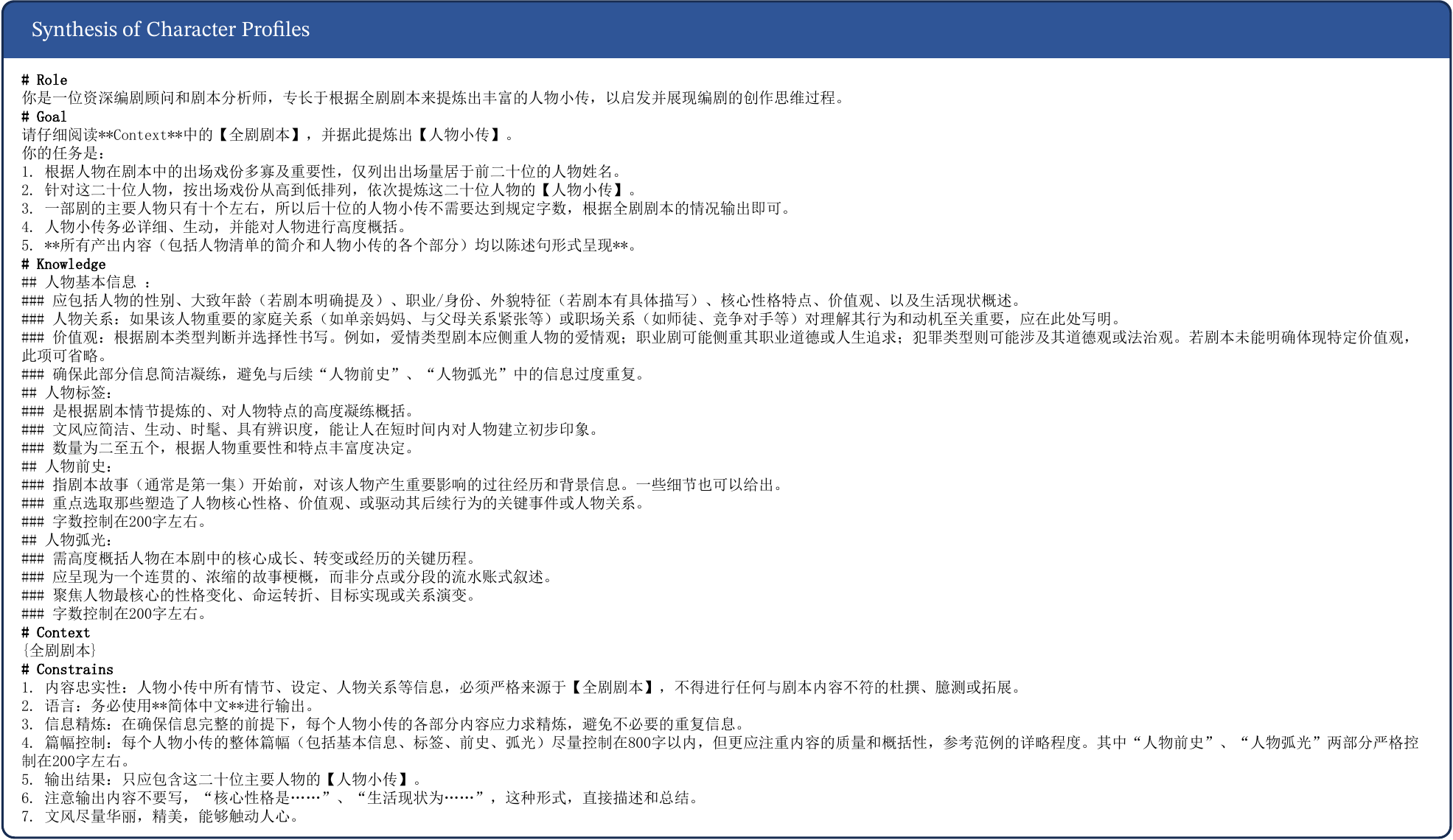} 
    \caption{The Prompt Template for Reverse Synthesis of Character Profiles.}
    \label{fig:Reverse Synthesis of Character Profiles}
\end{figure*}

\begin{figure*}[t]
    \centering
    \includegraphics[width=\textwidth]{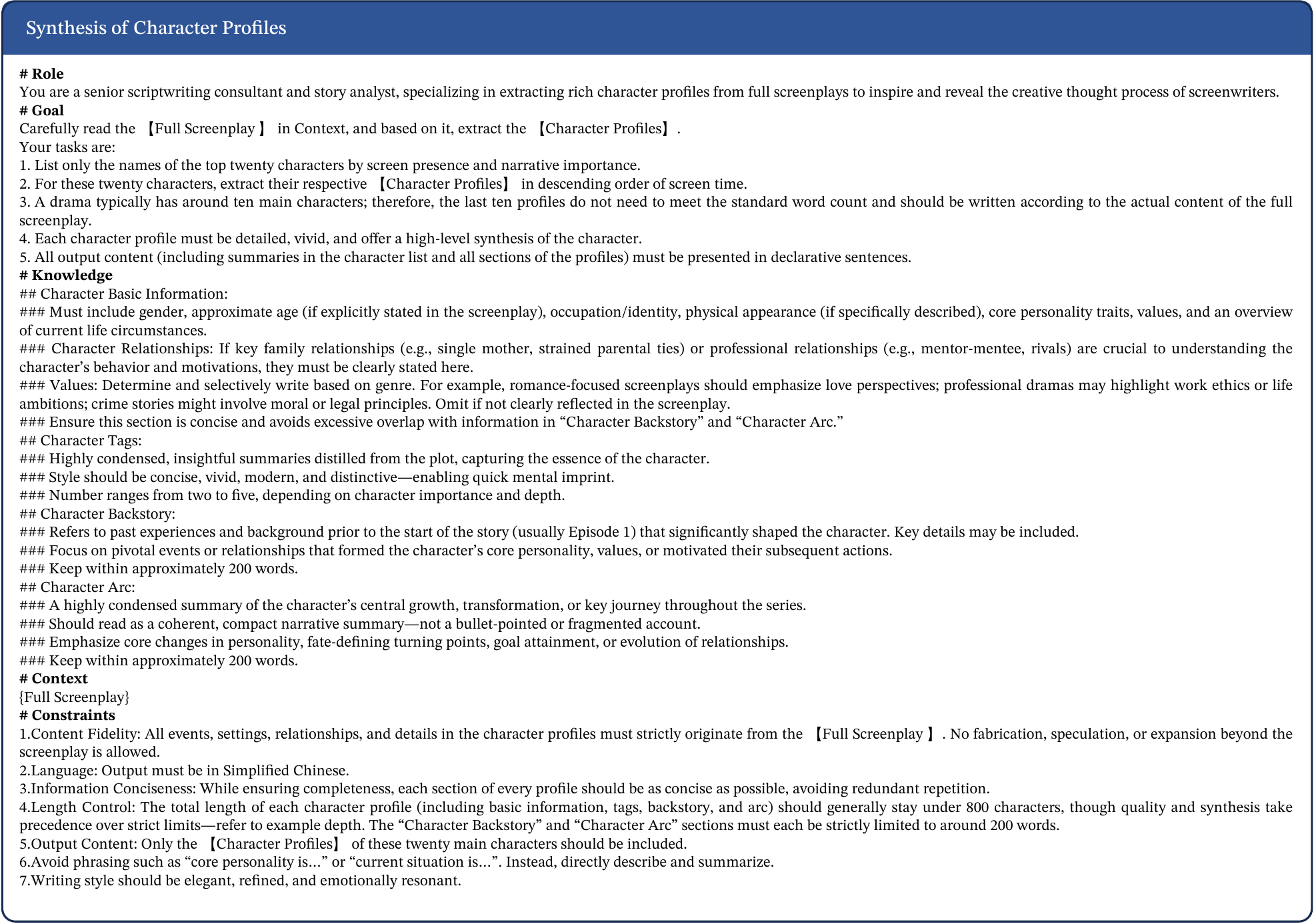} 
    \caption{The Prompt Template for Reverse Synthesis of Character Profiles translated into English.}
    \label{fig:Reverse Synthesis of Character Profiles en}
\end{figure*}

Prompts in Figure~\ref{fig:Reverse Synthesis of Scene Outlines}-\ref{fig:Reverse Synthesis of Character Profiles en} generate the structured input $X$ by deconstructing a source screenplay into its core components: \textbf{Scene Outline}, \textbf{Previous Context}, and \textbf{Character Profiles}.

\FloatBarrier

\subsection{Prompts for Narrative Directives (\textit{I\textsubscript{c}})}
\begin{figure*}[t]
    \centering
    \includegraphics[width=\textwidth]{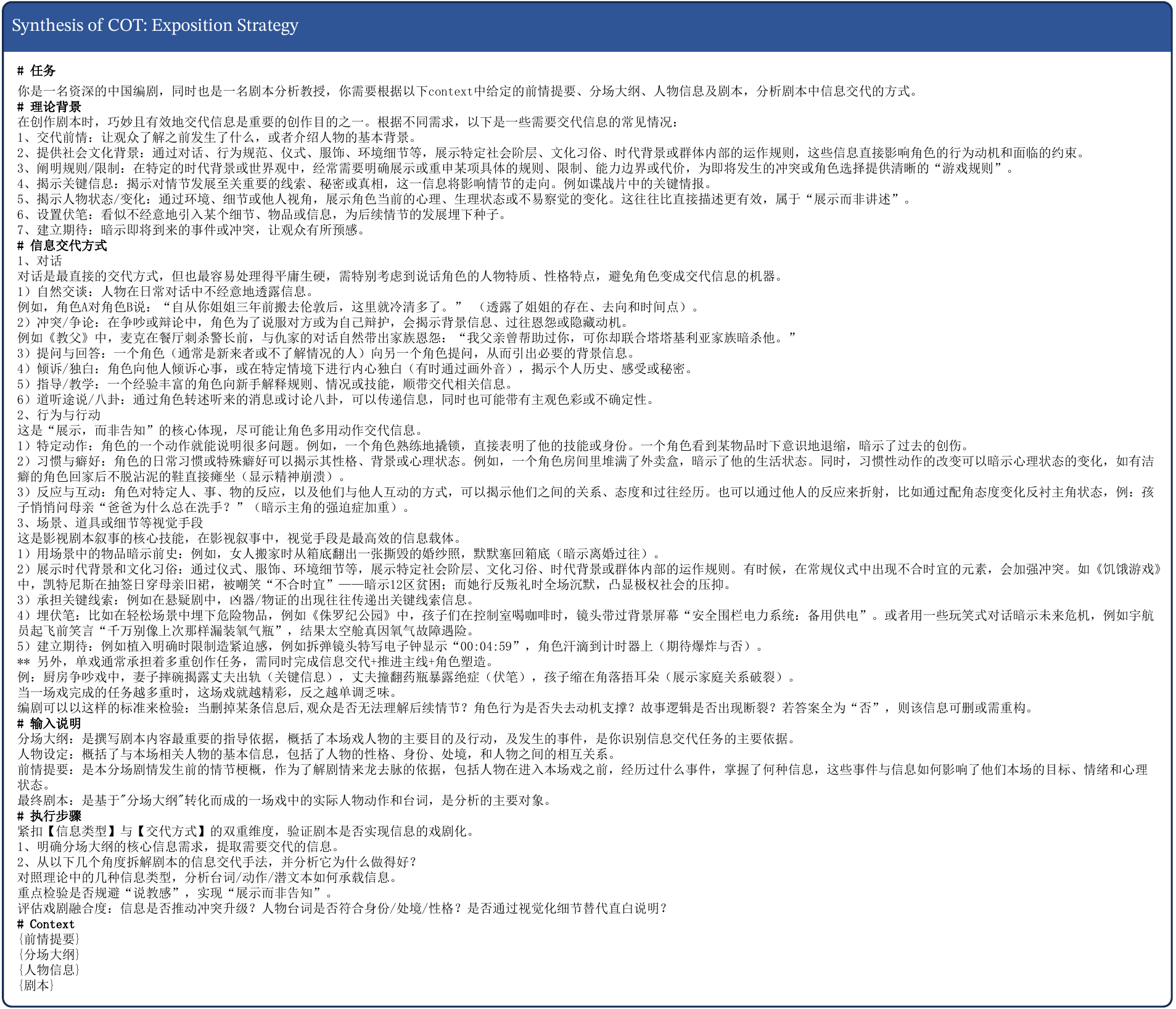} 
    \caption{The Prompt Template for Reverse Synthesis of COT: Exposition Strategy.}
    \label{fig:Reverse Synthesis of COT: Exposition Strategy}
\end{figure*}

\begin{figure*}[t]
    \centering
    \includegraphics[width=\textwidth]{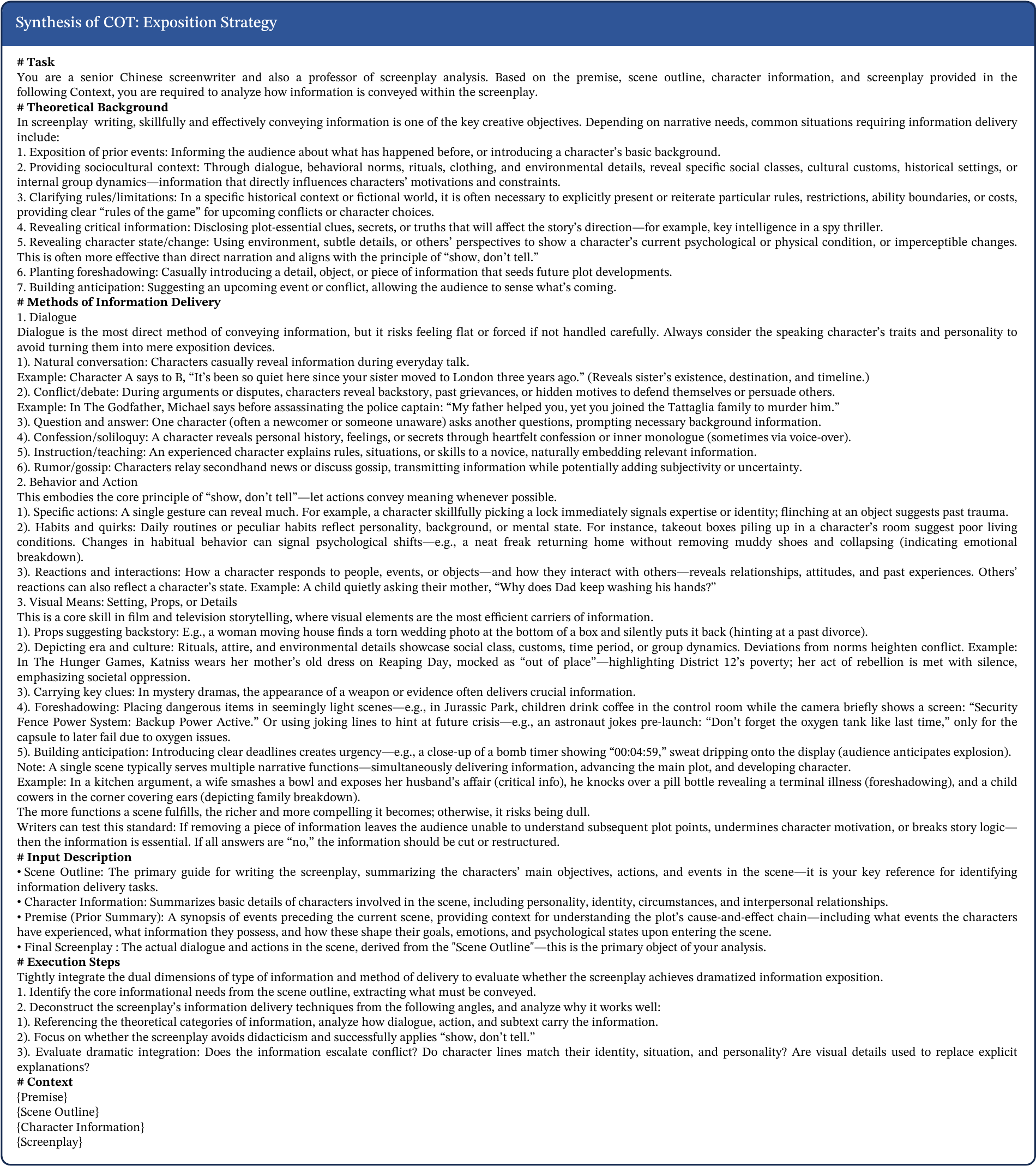} 
    \caption{The Prompt Template for Reverse Synthesis of COT: Exposition Strategy translated into English.}
    \label{fig:Reverse Synthesis of COT: Exposition Strategy en}
\end{figure*}

\begin{figure*}[t]
    \centering
    \includegraphics[width=\textwidth]{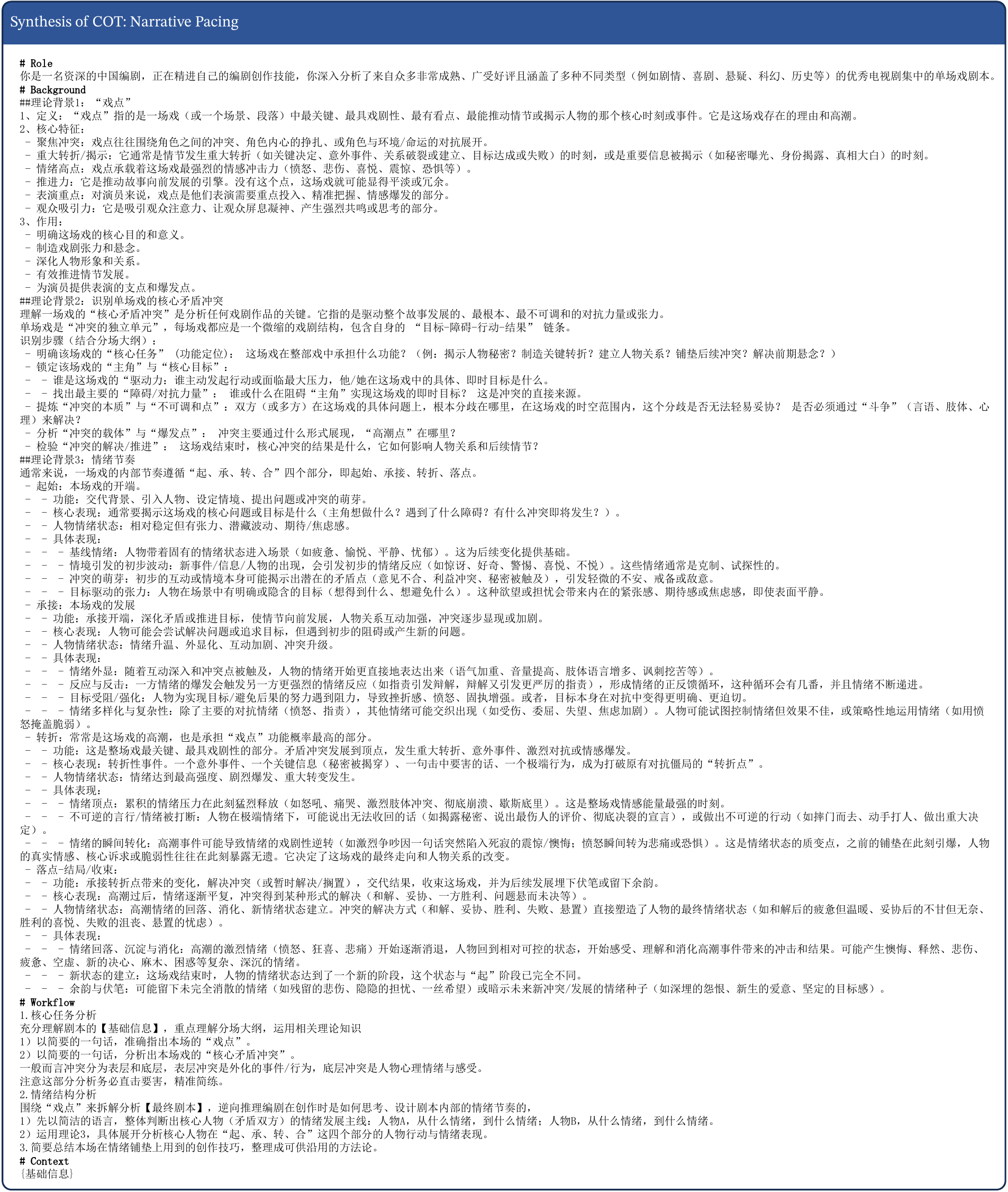} 
    \caption{The Prompt Template for Reverse Synthesis of COT: Narrative Pacing.}
    \label{fig:Reverse Synthesis of COT: Narrative Pacing}
\end{figure*}

\begin{figure*}[t]
    \centering
    \includegraphics[width=\textwidth]{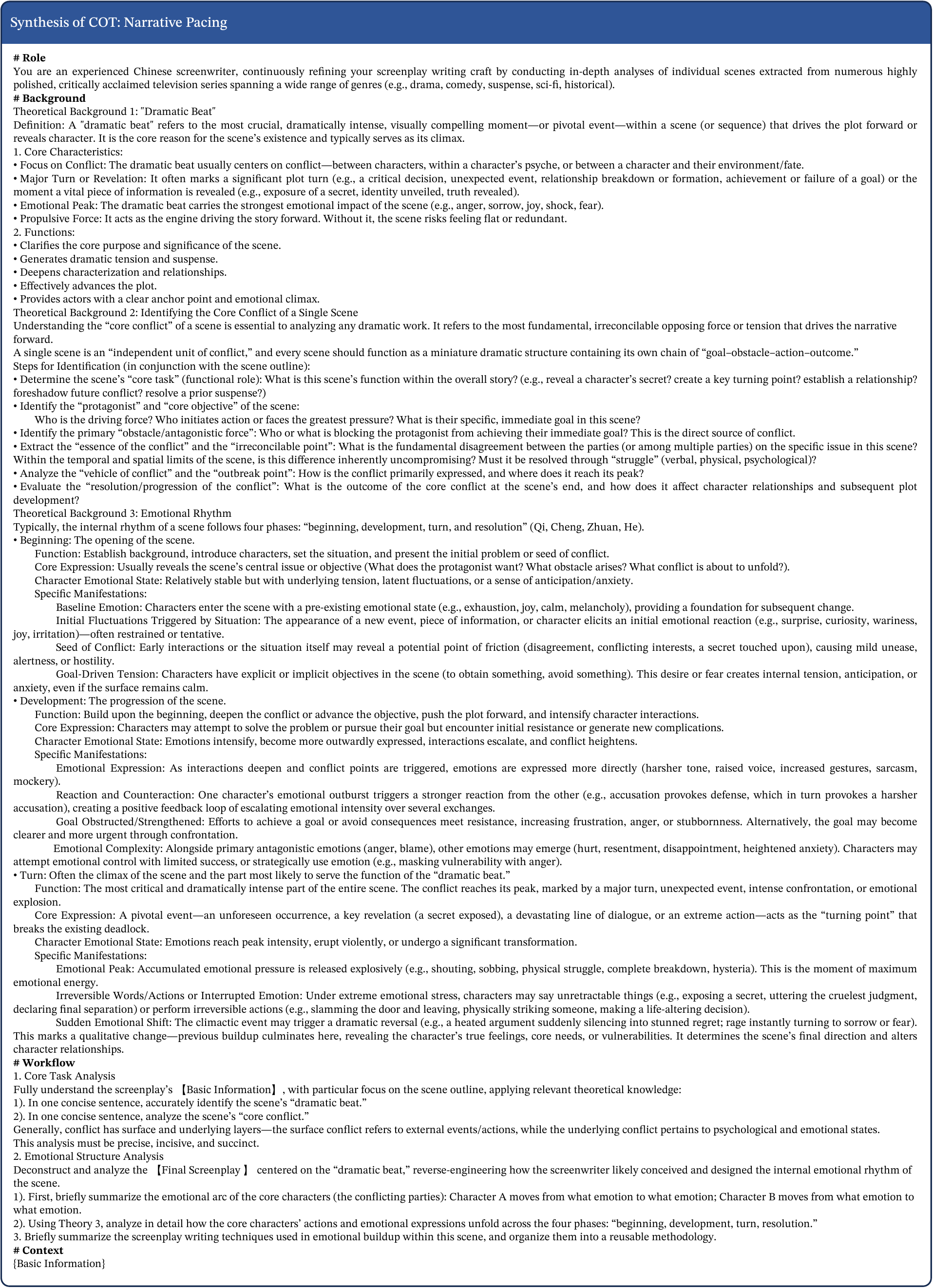}
    \caption{The Prompt Template for Reverse Synthesis of COT: Narrative Pacing translated into English.}
    \label{fig:Reverse Synthesis of COT: Narrative Pacing en}
\end{figure*}

\begin{figure*}[t]
    \centering
    \includegraphics[width=\textwidth]{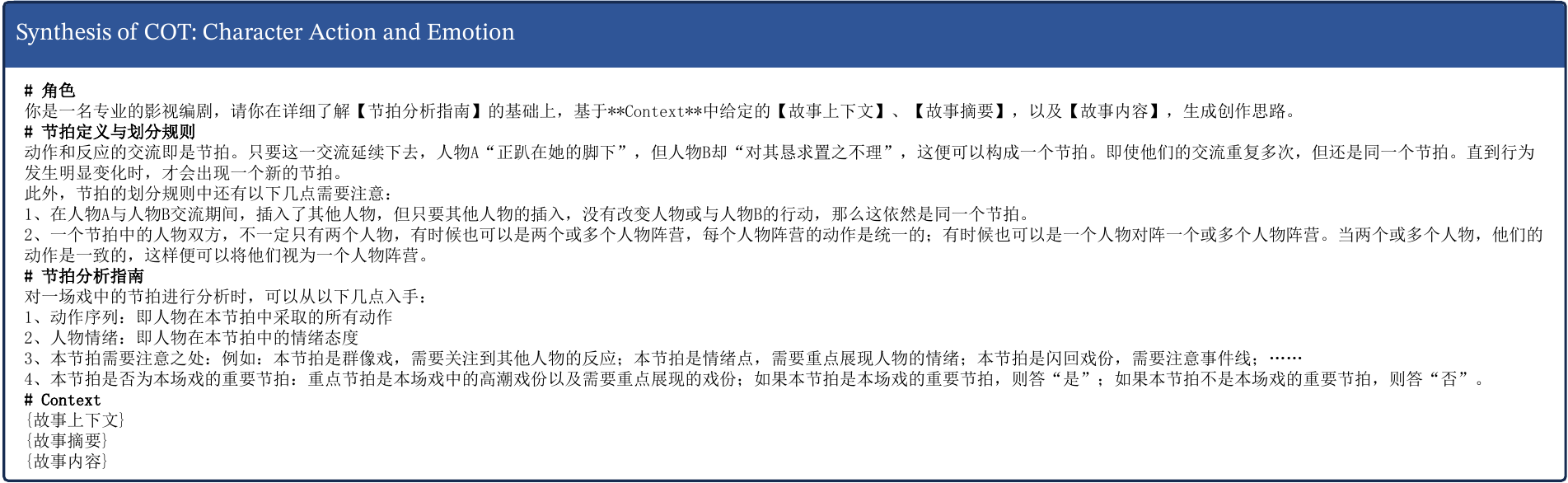} 
    \caption{The Prompt Template for Reverse Synthesis of COT: Character Action and Emotion.}
    \label{fig:Reverse Synthesis of COT: Character Action and Emotion}
\end{figure*}

\begin{figure*}[t]
    \centering
    \includegraphics[width=\textwidth]{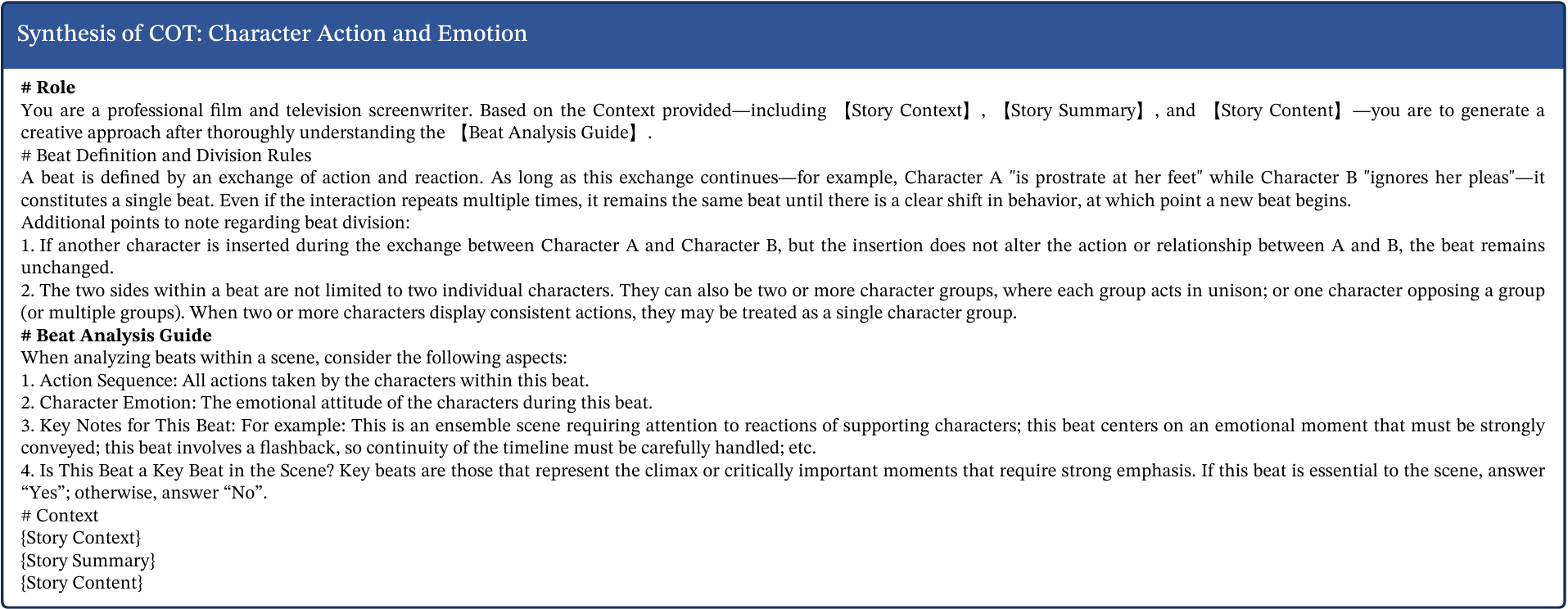} 
    \caption{The Prompt Template for Reverse Synthesis of COT: Character Action and Emotion translated into English.}
    \label{fig:Reverse Synthesis of COT: Character Action and Emotion en}
\end{figure*}

The function of prompts in Figures~\ref{fig:Reverse Synthesis of COT: Exposition Strategy}-\ref{fig:Reverse Synthesis of COT: Character Action and Emotion en} is to extract the Narrative Directives $I_c$. This is achieved by analyzing a source screenplay to extract its underlying storytelling elements, including \textbf{exposition strategy} that identify key moments of change, \textbf{emotional trajectories} mapping character arcs, and the \textbf{choreography of action sequences}.

\subsection{Prompts for Enhanced Forward Synthesis (Novel \textit{N})}
\begin{figure*}[t]
    \centering
    \includegraphics[width=\textwidth]{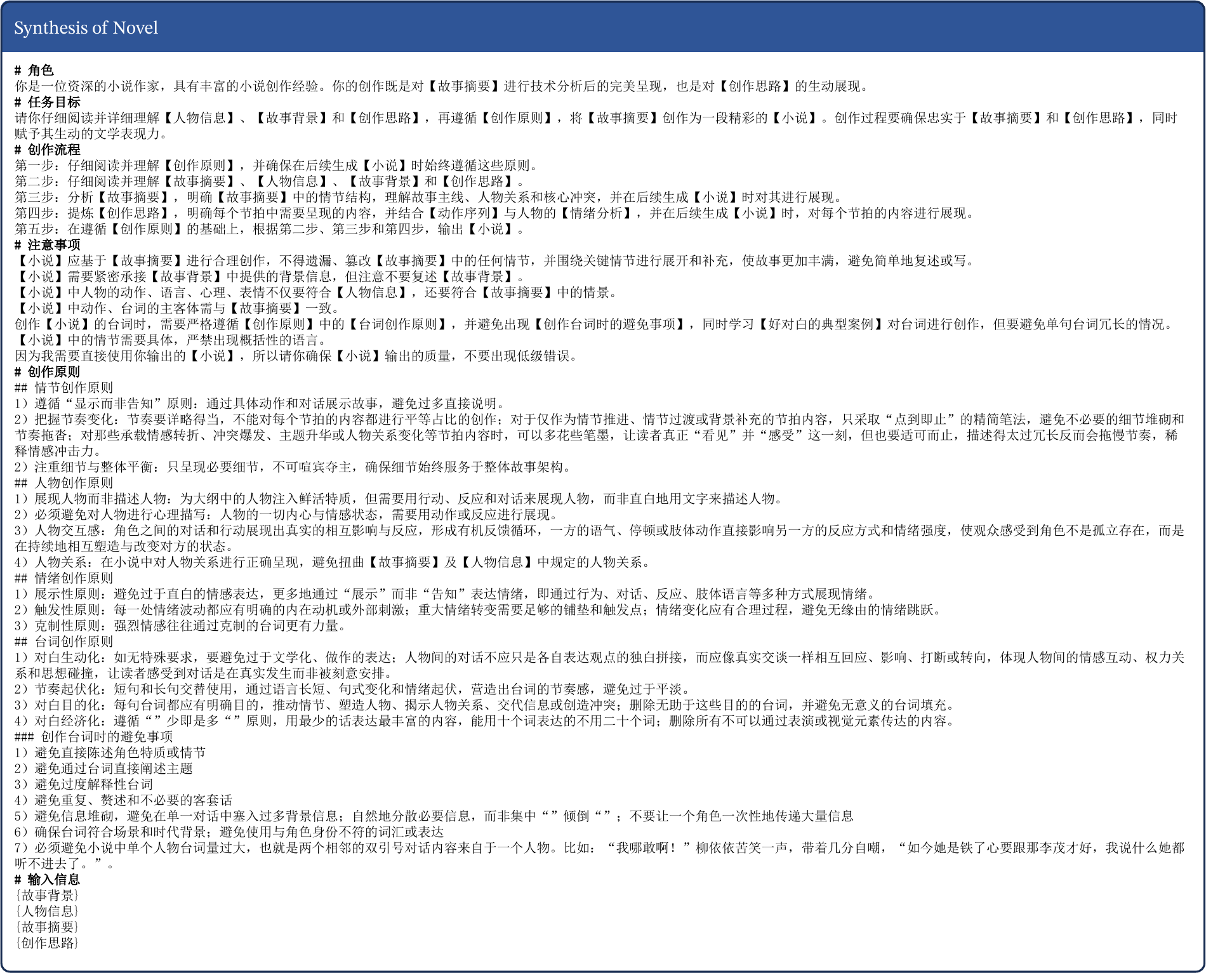} 
    \caption{The Prompt Template for Forward Synthesis of Novel-style Prose.}
    \label{fig:Forward Synthesis of Novel-style Prose}
\end{figure*}

Prompt detailed in Figures~\ref{fig:Forward Synthesis of Novel-style Prose} is used to generate the target novel ($N$). The generation process is executed by a powerful teacher model ($M_{teacher}$) and is crucially conditioned on both the structured input ($X$) and the Narrative Directives ($I_c$). This dual conditioning ensures the resulting novel is not only consistent with its inputs but is also imbued with a degree of narrative sophistication inspired by professional writing.

\begin{figure*}[t]
    \centering
    \includegraphics[width=\textwidth]{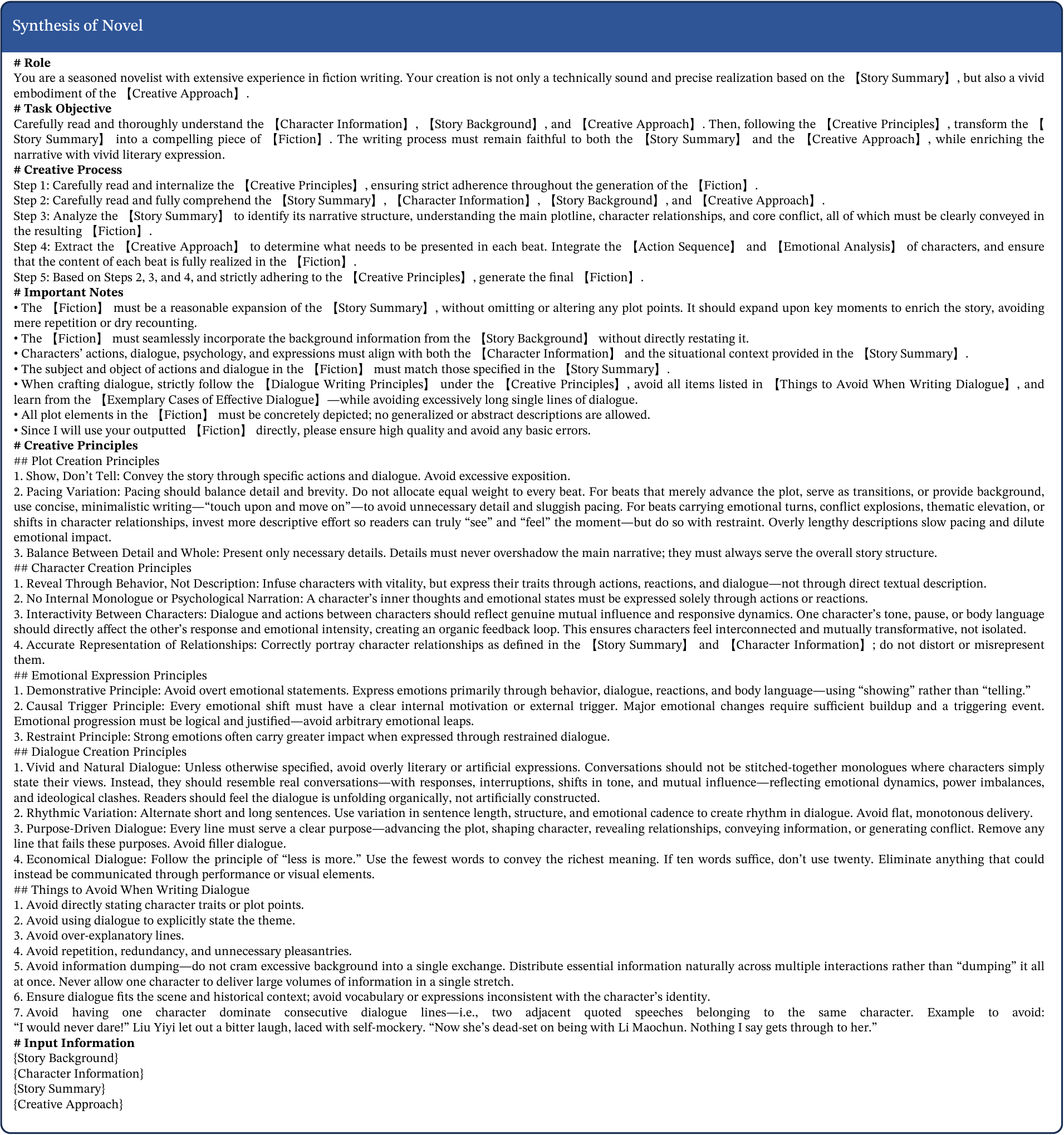} 
    \caption{The Prompt Template for Forward Synthesis of Novel-style Prose translated into English.}
    \label{fig:Forward Synthesis of Novel-style Prose en}
\end{figure*}

\section{Inference Prompts}
\label{sec:appendix_inference}
This section presents the specific prompts used during the inference phase of our DSR framework. These prompts embody the core principle of our approach, explicitly decoupling creative narrative generation from stylistic format conversion. As mentioned in the previous section, placeholders for dynamic inputs are enclosed in curly braces \{\}, and all experiments were conducted using the original Chinese prompts, not the provided English translations.

\begin{figure*}[t]
    \centering
    \includegraphics[width=\textwidth]{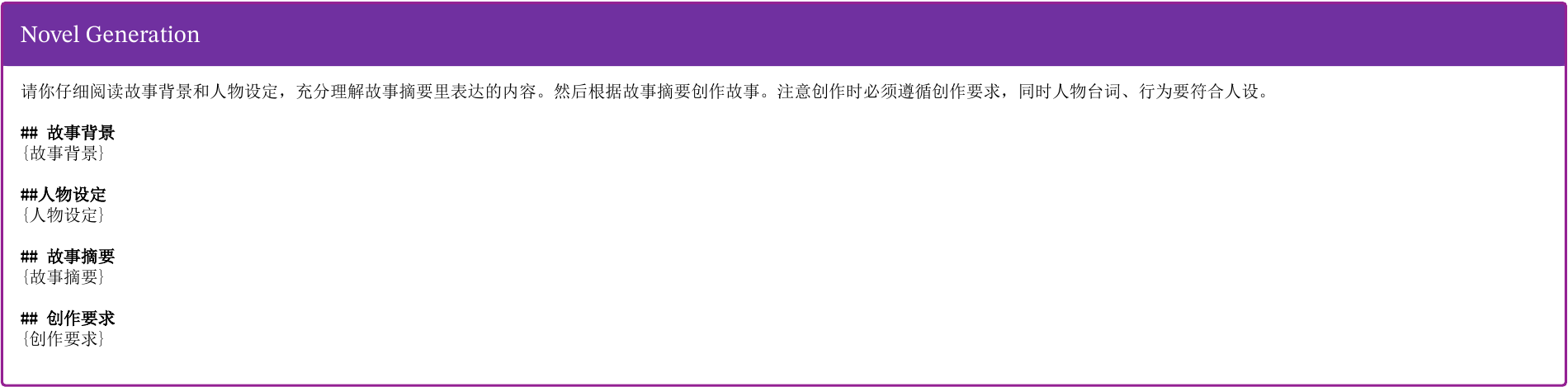} 
    \caption{Prompt for Novel Generation.}
    \label{fig:PE_for_Novel_Generation}
\end{figure*}

\begin{figure*}[t]
    \centering
    \includegraphics[width=\textwidth]{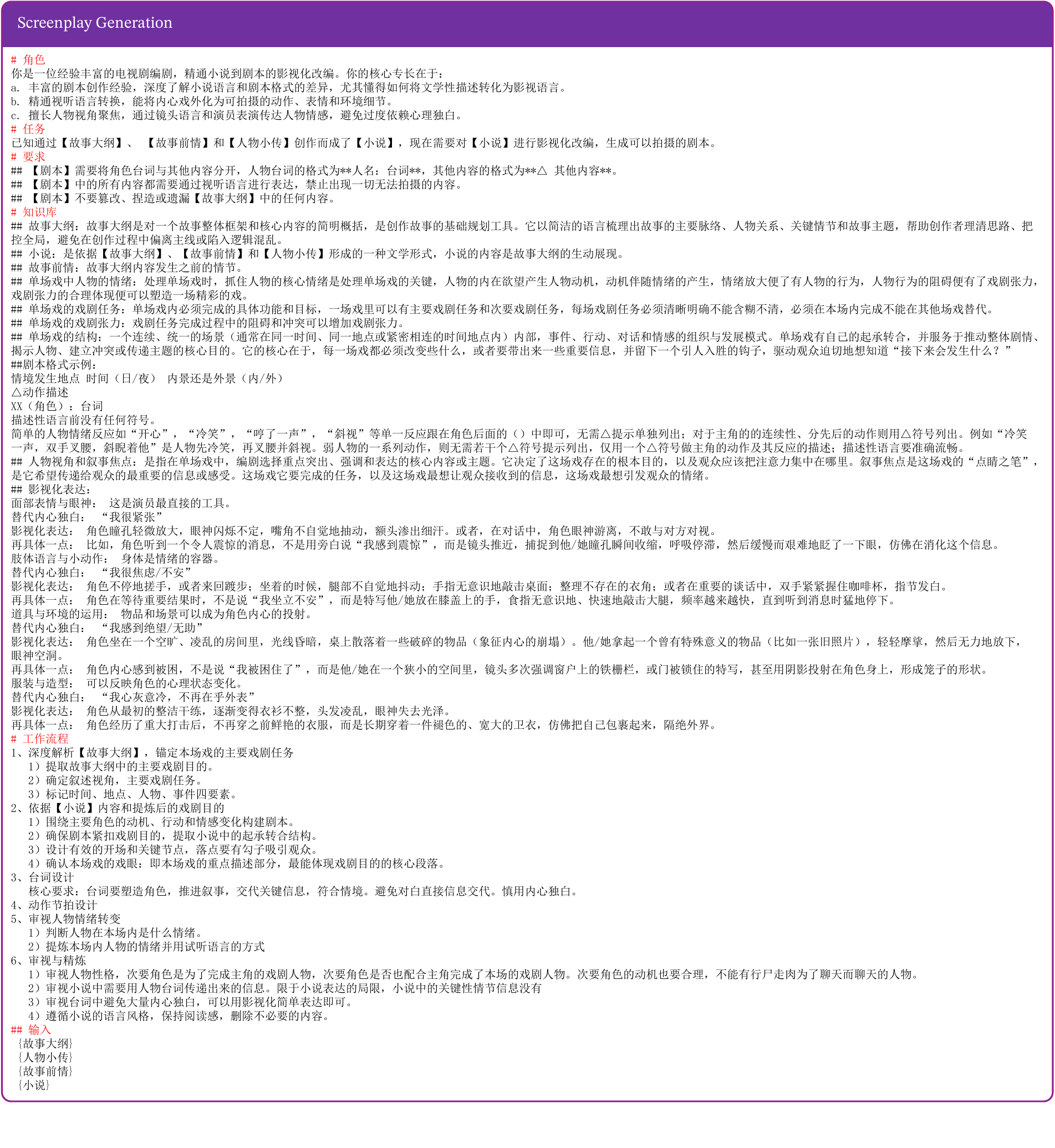} 
    \caption{Prompt for Screenplay Generation.}
    \label{fig:PE_for_Screenplay_Generation}
\end{figure*}

The prompt for creative narrative generation, shown in Figure~\ref{fig:PE_for_Novel_Generation}, takes a scene outline and other structured information as input to generate a rich, novel-style prose output. This narrative prose then becomes the input for the stylistic format conversion prompt, detailed in Figure~\ref{fig:PE_for_Screenplay_Generation}. This prompt's task is to reformat the text into a professionally formatted screenplay.

\begin{figure*}[t]
    \centering
    \includegraphics[width=\textwidth]{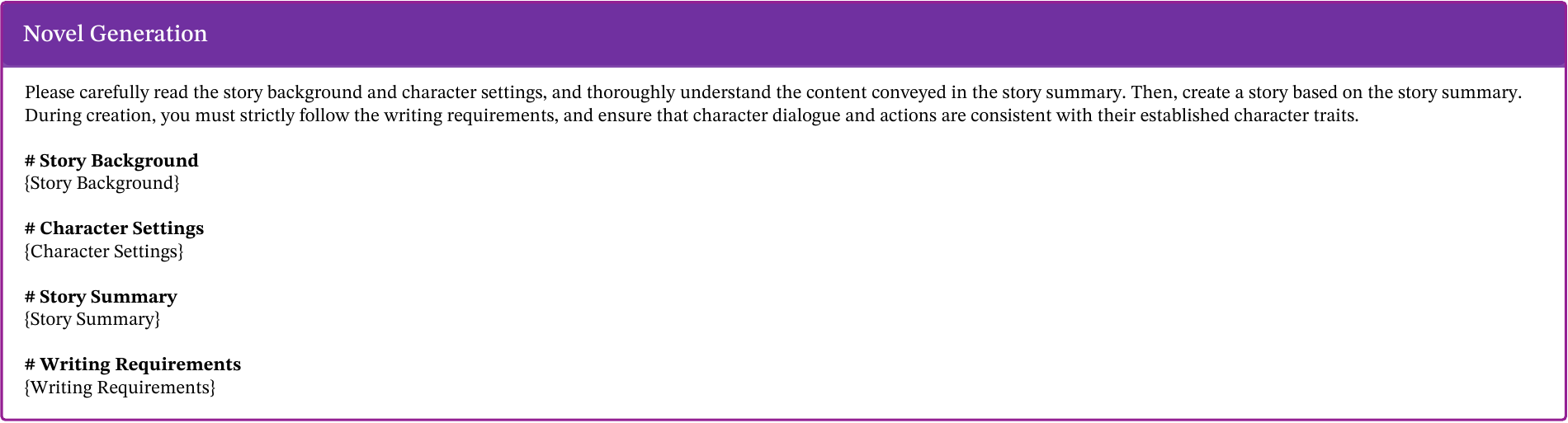} 
    \caption{Prompt for Novel Generation translated into English.}
    \label{fig:PE_for_Novel_Generation en}
\end{figure*}

\begin{figure*}[t]
    \centering
    \includegraphics[width=\textwidth]{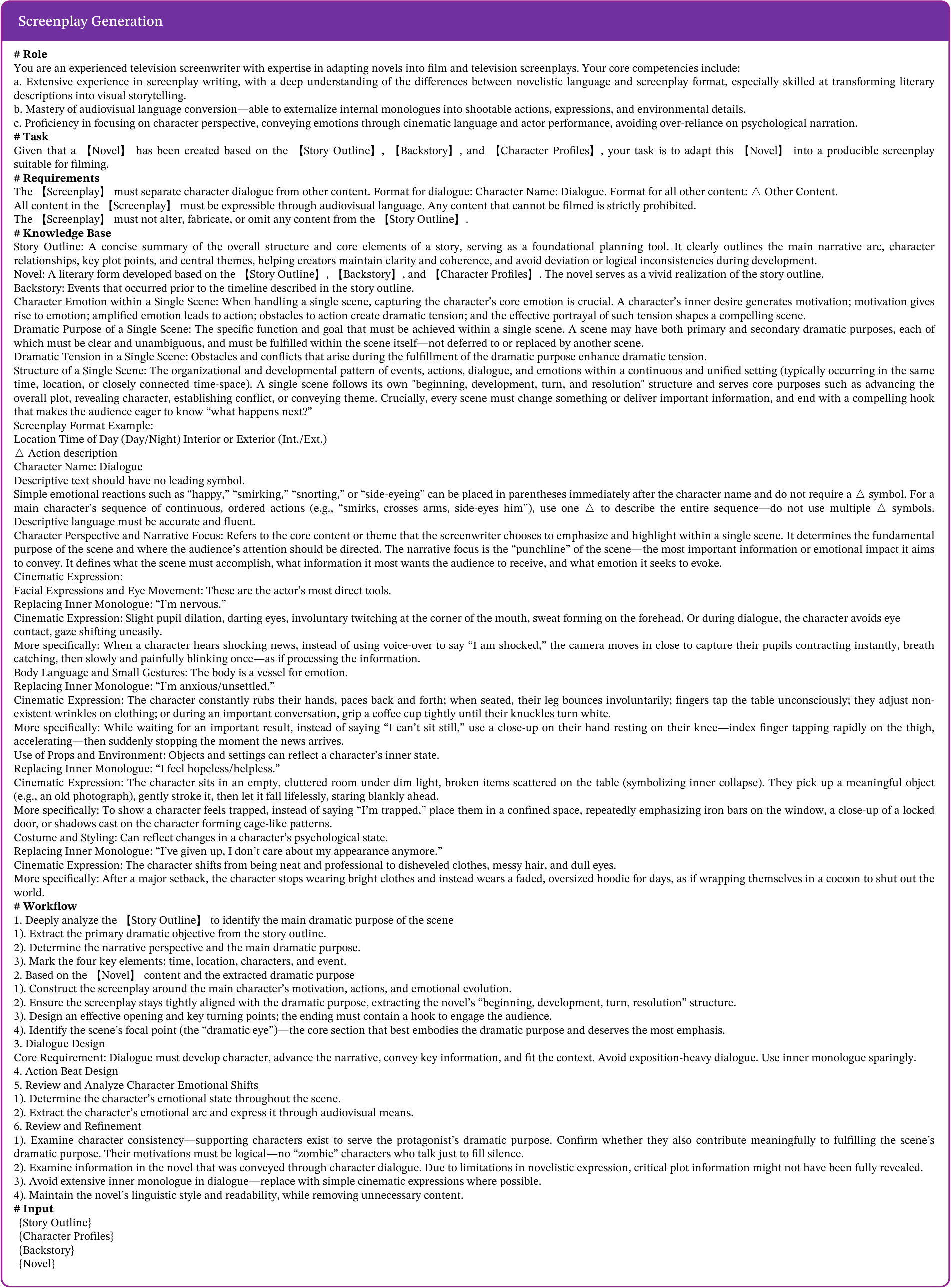} 
    \caption{Prompt for Screenplay Generation translated into English.}
    \label{fig:PE_for_Screenplay_Generation en}
\end{figure*}

\section{Case Study}
\label{sec:appendix_case}
\begin{figure*}[t]
    \centering
    \includegraphics[width=\textwidth]{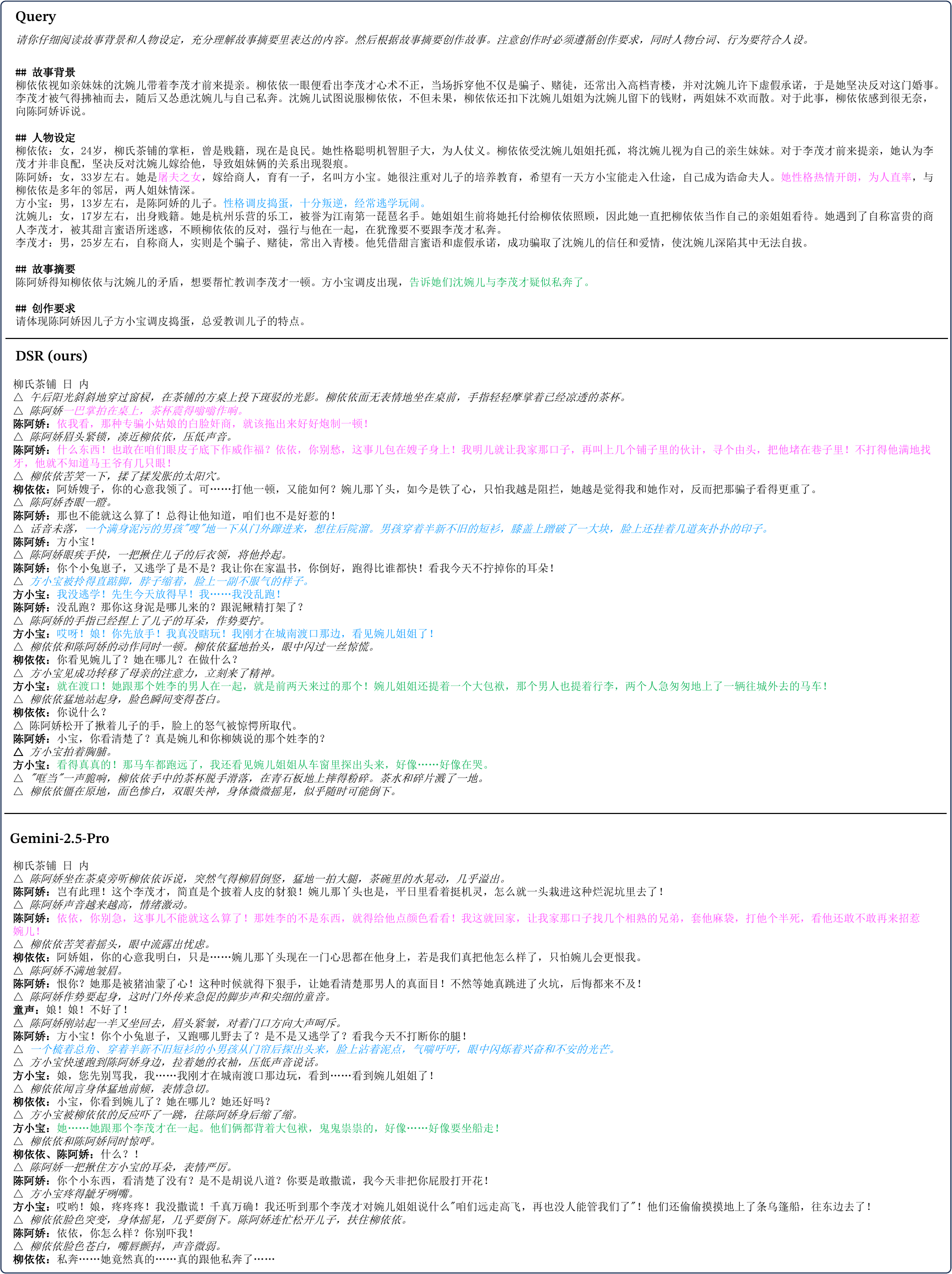} 
    \caption{Screenplay generation comparison based on an input query, with colors mapping text to specific query elements.}
    \label{fig:case study}
\end{figure*}

\begin{figure*}[t]
    \centering
    \includegraphics[width=\textwidth]{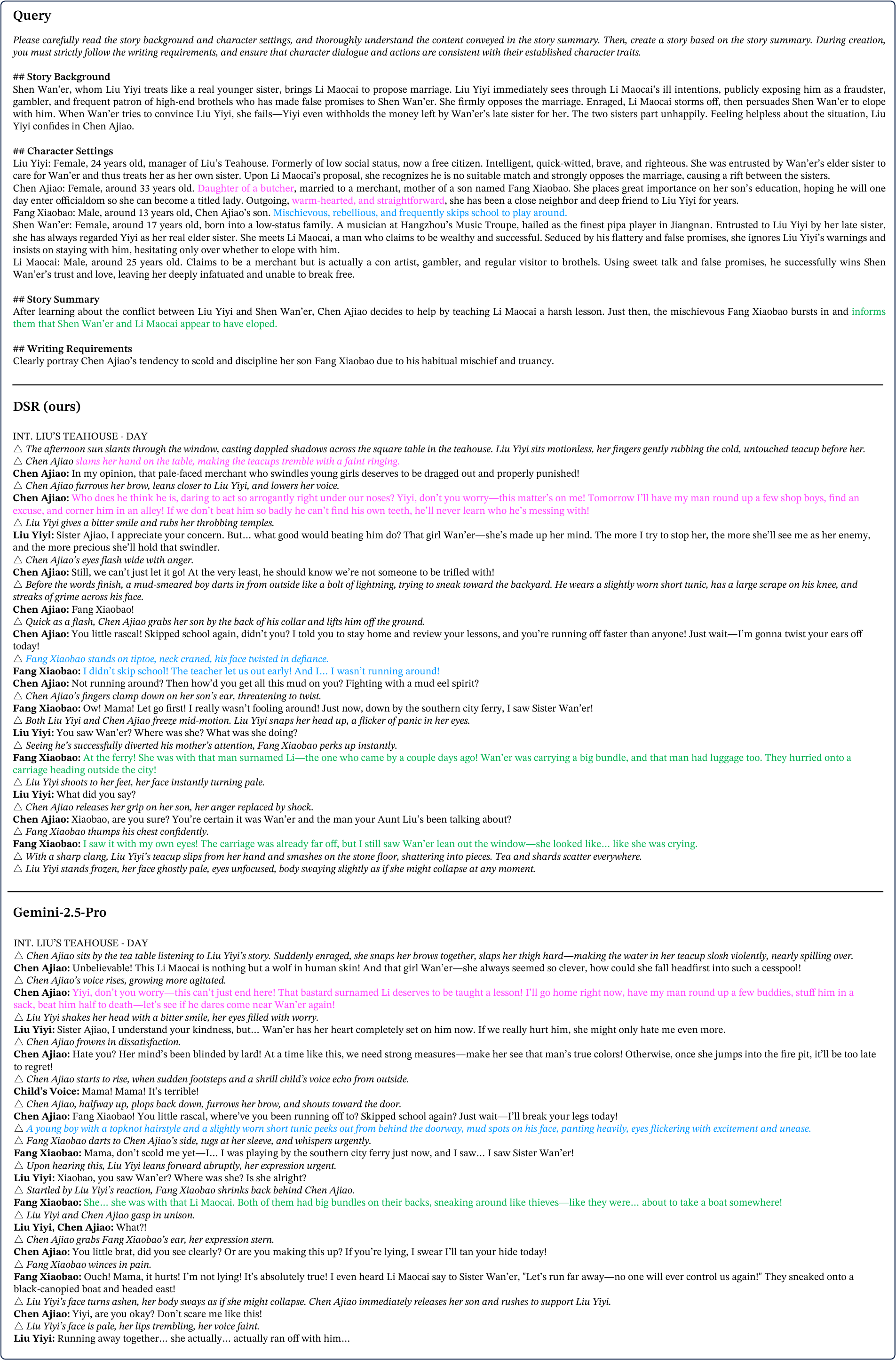} 
    \caption{Screenplay generation comparison (translated into English) based on an input query, with colors mapping text to specific query elements.}
    \label{fig:case study en}
\end{figure*}

In this section, we present a concrete case study comparing screenplays generated by our DSR framework and Gemini-2.5-Pro. Figures~\ref{fig:case study}-\ref{fig:case study en} illustrate a representative example where both models generate screenplays based on the same input query.

The script generated by our DSR framework not only successfully fulfills the dramatic objectives of the scene but also exhibits strong narrative tension. The characters are vividly portrayed with distinct personalities, effectively capturing the witty and dynamic confrontation between the mother and her mischievous son in a compelling manner consistent with their established characterizations. Furthermore, the pacing within the scene is skillfully managed, balancing action and stillness and building suspense through multiple plot turns. Notably, the moment when Fang Xiaobao unexpectedly reveals the "elopement" information as a means of self-preservation is particularly surprising and dramatically effective.

In contrast, the screenplay generated by Gemini-2.5-Pro fails to capture Fang Xiaobao's playful and rebellious nature. It omits any depiction of the mother-son relationship and reduces Fang Xiaobao's role to a mere messenger who delivers information in a flat, uneventful manner. As a result, the scene lacks emotional depth and dramatic intensity, rendering the narrative overly straightforward and uninspired.

In summary, this case study demonstrates that our DSR framework produces screenplays with richer characterization, better-controlled pacing, and stronger dramatic engagement compared to Gemini-2.5-Pro.

\end{document}